\documentclass[smallextended]{svjour3}       
\smartqed  
\usepackage{graphicx}
\usepackage{times}
\usepackage{epsfig}
\usepackage{amsmath}
\usepackage{amssymb}
\usepackage{gensymb}
\usepackage{url}

\usepackage{array}
\usepackage{multirow}
\usepackage{textcomp}
\usepackage{array}
\usepackage{colortbl}
\usepackage[table]{xcolor}
\usepackage{booktabs}

\definecolor{myblue}{rgb}{0,0,1}
\definecolor{myred}{rgb}{0.8, 0, 0}
\definecolor{mygreen}{rgb}{0, 0.6, 0}

\newcommand{\ie}{{\it i.e.},~}
\newcommand{\etal}{{\it et al.}~}

\begin{document}

\title{Post-mortem Iris Decomposition and its Dynamics in Morgue Conditions}

\author{Mateusz Trokielewicz \and Adam Czajka \and Piotr Maciejewicz}

\institute{{\bf Mateusz Trokielewicz} \at
Warsaw University of Technology \\
Institute of Control and Computation Engineering\\
Nowowiejska 15/19, 00665 Warsaw, Poland\\
\email{m.trokielewicz@elka.pw.edu.pl}\\
https://orcid.org/0000-0002-7363-8385           
\and
{\bf Adam Czajka} \at
Department of Computer Science and Engineering\\
University of Notre Dame,\\
Notre Dame, IN 46556, USA\\
\email{aczajka@nd.edu}\\
https://orcid.org/0000-0003-2379-2533           
\and
{\bf Piotr Maciejewicz} \at
Department of Ophthalmology\\
Medical University of Warsaw\\
Lindleya 4, 02005 Warsaw, Poland\\
\email{piotr.maciejewicz@wum.edu.pl}\\
https://orcid.org/0000-0001-6725-6332
}

\date{Received: date / Accepted: date}

\maketitle
\begin{abstract}
With increasing interest in employing iris biometrics as a forensic tool for identification by investigation authorities, there is a need for a thorough examination and understanding of post-mortem decomposition processes that take place within the human eyeball, especially the iris. This can prove useful for fast and accurate matching of ante-mortem with post-mortem data acquired at crime scenes or mass casualties, as well as for ensuring correct dispatching of bodies from the incident scene to a mortuary or funeral homes. Following these needs of forensic community, this paper offers an analysis of the coarse effects of eyeball decay done from a perspective of automatic iris recognition point of view. Therefore, we analyze post-mortem iris images acquired in both visible light as well as in near-infrared light (860 nm), as the latter wavelength is used in commercial iris recognition systems. Conclusions and suggestions are provided that may aid forensic examiners in successfully utilizing iris patterns in post-mortem identification of deceased subjects. Initial guidelines regarding the imaging process, types of illumination, resolution are also given, together with expectations with respect to the iris features decomposition rates.
\end{abstract}

\section{Introduction }
\label{sec:Intro}
\subsection{Post-mortem iris biometrics}
DNA, fingerprints, facial characteristics, or dental records have been widely used in forensic science and practice for identification of victims. It has been recently shown that automatic iris recognition methods are able to offer perfect accuracy on identification of samples acquired 5-7 hours post-mortem, and when the bodies were kept in low temperatures, correct matches were occasionally possible for images collected up until approximately three weeks after death \cite{TrokielewiczPostMortemBTAS2016,BolmeBTAS2016,Sauerwein_JFO_2017,TrokielewiczTIFS2018}. It can be assumed that such three-week periods are long enough for some forensic investigations and usage of regular iris recognition matchers may suffice. We believe that simultaneous investigation of post-mortem iris changes from medical and engineering standpoints may facilitate construction of specialized iris recognition methods that are better suited to forensic tasks.

\subsection{Motivation}
There are a few scenarios in which post-mortem iris recognition can be useful due to its speed (when compared to usually slower DNA analysis), and which have been already appreciated by forensics community.

One of them is more accurate matching of ante-mortem samples (if available) with post-mortem data obtained at crime scenes and mass fatality incidents. One of the United States Department of Justice's operational requirements is defined as: ``Enhancement of unidentified decedent system(s) with weighting capability for antemortem and postmortem comparisons with the goal of providing a ranked list of “best matches” to effectively and efficiently identify potential candidates or hits.''\footnote{\url{https://www.nij.gov/topics/forensics/documents/2018-2-forensic-twg-table.pdf}}.

The second, operational practice-driven application is to rapidly register the body at the scene, to track it later and dispatch it correctly to family or mortuary. In case of mass fatality accidents, DNA is too slow and forensic practitioners already decided to use iris recognition, which has been demonstrated for the first time in 2015 in Sansola's master thesis defended at the Boston Medical School\footnote{\url{https://open.bu.edu/handle/2144/13975}}.

Post-mortem iris recognition also received an attention of the National Institute of Standards and Technology -- the organizer of the Iris Expert Group meetings addressing important iris recognition challenges from the US government point of view. The post-mortem iris recognition has been discussed at the IEG meetings in 2016, 2018 and 2019\footnote{\url{https://www.nist.gov/programs-projects/iris-experts-group-ii-homepage}}. 

\subsection{Paper contributions}
The purpose of this paper is to provide a thorough analysis of the processes that take place in the human eye after death, their dynamics, and influence on the appearance of the iris from an automatic biometric identification point of view. We describe post-mortem global changes observed in cadavers' eyes, and present feature-level close-up analysis of selected image patches that may contribute to training specialized feature extraction algorithms. This is done for the mortuary conditions, where cadavers reside in low temperatures.

To our knowledge, this is the first work thoroughly analyzing the decomposition of both the eyeball as a whole, as well as individual iris features that are employed for identification in iris recognition systems. We hope that conclusions made by engineers having an expertise in automatic iris recognition would be potentially useful for forensic examiners in their practice when dealing with post-mortem iris data processing.

The paper is laid out as follows. Section \ref{sec:Related} reviews the appropriate scientific literature related to post-mortem iris recognition. Section \ref{sec:Medical} provides a medical background regarding post-mortem changes to the eye tissues, mainly from the biometric standpoint. Section \ref{sec:Database} briefly describes the dataset used for these experiments (offered along with the paper). Section \ref{sec:Global} contains an analysis of post-mortem changes to the iris together with the qualitative assessment of their dynamics. Close-up analysis on the decomposition of iris features is then provided in Section \ref{sec:CloseUp}, together with manual annotations of individual features in the iris throughout the decomposition timeline performed by the Authors. Relevant conclusions are delivered in Section \ref{sec:Conclusions}.

\section{Related work }
\label{sec:Related}
\subsection{Post-mortem iris recognition feasibility studies}
Several scientific studies aimed at quantifying the biometric performance of post-mortem iris samples with existing iris recognition technologies. The first was Sansola, who experimented on data collected from 43 subjects and had iris image samples collected at different intervals after death using IriTech's M2120U sensor, and then enrolled in the IriCore iris matching software \cite{BostonPostMortem}. The post-mortem verification yielded approximately 30\% of false non-matches with 0\% false matches.

Saripalle \etal studied post-mortem iris recognition possibilities with ex-vivo eyes of a domestic pig \cite{PostMortemPigs}, concluding that eyes taken out of the body and stored in room temperature are rapidly degrading and lose their biometric capability after 6 to 8 hours.

Ross \cite{RossPostMortem} presented a few observations related to the fadeout of the pupillary and limbic boundaries, and corneal opacity found in some post-mortem iris images.

Trokielewicz \etal have experimentally shown that the human iris does not immediately degrade upon death \cite{TrokielewiczPostMortemICB2016},\cite{TrokielewiczPostMortemBTAS2016}. They concluded that iris pattern was visible up to a few days after demise, with pupils usually fixed in the so-called 'cadaveric position' with a neutral dilation. After a few hours after death, 90\%-100\% of their samples were still recognizable, depending on the iris recognition method employed. After a day, when the cornea started to become opaque, the recognition rate dropped to 13\%-73\%, with the best accuracy (73\%) achieved for the IriCore method (developed by the sensor's manufacturer). A long-term analysis of recognition accuracy was also carried out, with correct matches occasionally reported for samples collected as long as 17 days after death.

Bolme \etal tracked various biometric characteristics during in-the-wild decomposition \cite{BolmeBTAS2016}, including face, fingerprint, and iris. The first two were rather resilient to the outside conditions and passing time, being able to provide biometric traits even after many days of subject placement. The irises, however, were degrading very quickly in their experiments, regardless of the season of the year and temperature outside. The authors state that irises typically became useless from the recognition viewpoint only a few days after placement, allowing correct verification rate of only 0.6\% with samples collected up to 14 days after exposing the subject to outdoor conditions (yet there were no details revealed on whether the correctly matched samples are those collected shortly after the placement, or 14 days after the placement.) The real-life chance of recognizing an iris was estimated as below 0.1\% by these authors.

In turn, Sauerwein \etal \cite{Sauerwein_JFO_2017} showed that irises stay readable for up to 34 days after death, when cadavers were kept in outdoor conditions during winter. This readability parameter, however, was assessed by human experts acquiring the samples and no automatic iris recognition algorithms were used in this study.

More recently, Trokielewicz \etal \cite{TrokielewiczTIFS2018} involved four iris recognition methods to analyze the genuine and impostor comparison scores and evaluate the dynamics of iris decay over a period of up to 814 hours. The findings show that iris recognition may be close-to-perfect 5 to 7 hours after death and occasionally viable even 21 days after death, and that false match probability is significantly higher when live iris images are compared with post-mortem samples than when only live samples are used in comparisons.

\subsection{Methods designed for post-mortem iris biometrics}
Some attempts have been made in respect to improving the iris recognition accuracy in a post-mortem scenario. Trokielewicz \etal proposed a post-mortem iris image segmentation method based on convolutional neural networks, which achieves segmentation accuracy superior to an exiting open-source iris recognition method OSIRIS \cite{TrokielewiczIWBF2018}. This iris image segmentation model was further re-trained with new data and deployed as part of a standard, iris code-based recognition pipeline \cite{DaugmanPatent}, which allowed the authors to obtain post-mortem matching accuracy that is superior to both the open-source as well as commercial, off-the-shelf algorithms \cite{TrokielewiczIMAVISColdSeg2019}.

A presentation attack detection method designed to discern live iris presentations to the biometric sensor from those where cadaver tissue is used has also been proposed, reaching 1\% of attack presentation classification error rate (\ie chance of a successful presentation attack) while maintaining 0\% bona-fide presentation classification error rate (\ie a chance of rejecting an authentic eye) \cite{TrokielewiczColdPAD_BTAS2018}.

Although a considerable number of studies regarding the \emph{automatic} performance of post-mortem iris recognition has been carried out, only a few included a \emph{human perception} factor, such as a study by Czajka \etal \cite{CzajkaBSIFIrisDesc2019}, who proposed an iris-specific, human-driven iris recognition method. In their experiment, humans were presented iris image pairs (including post-mortem samples) and were asked to (1) say if the pair shows the same eye or different eyes, and (2) annotate regions that supported their decision. Additionally, the subjects' gaze was recorded by an eye tracking device. Salient regions either annotated by humans, or concluded from an eye tracking records, were then used to develop filtering kernels extracting features that -- in hope of the authors -- were closer to features used by humans in a similar experiment. Indeed, their method achieved results that were superior to a few selected open-source iris recognition tools.

A PhD dissertation by Trokielewicz introduces the first iris feature representation designed specifically to deal with post-mortem samples \cite{Trokielewicz_PhD_2019}. By utilizing data-driven filter kernels trained with Siamese networks on post-mortem iris data together with a state-of-the-art iris segmentation solution also devised specifically for the cadaver samples, an iris recognition method outperforming the gold standard commercial iris matchers is described. 

Czajka \etal \cite{Czajka_BTAS_2019} compared humans and a computer algorithm in their ``perception'' of salient iris features when solving the post-mortem iris recognition task. In the experiment employing an eye tracker device, human attention maps were confronted with attention maps obtained from a deep learning-based iris classifier. These studies allowed to formulate interesting conclusions, namely that the machine classifier can serve as a complementary tool to the forensic expert's knowledge, since combining its decisions with those provided by humans allowed for a significant increase in post-mortem iris verification accuracy.

\section{Medical background }
\label{sec:Medical}
Two biological, post-mortem processes set the direction of decay changes in a cadaver eye: autolysis (self-destruction) of the cells, and putrefaction (tissue decomposition). The former originates in hydrolytic enzymes released from the cells after demise, the latter is caused by microorganism activity.

The first tissue to be visibly affected by post-mortem decomposition is the cornea -- the outermost piece of the eyeball, normally kept in a hydrated state by tear film and blinking, it starts to slowly dry out after death, which leads to opacification, as the lacrimal glands producing the tear film no longer operate. This reduces the transparency of the cornea, with the rate of these processes being heavily dependent on the surrounding temperature, as well as on humidity, air movement, or closure of the eyelids \cite{Prieto2015}.

As for the iris itself, no evident changes to its tissues can be immediately spotted. No excessive constriction or dilation of the pupil is to be expected, as they are usually fixed in a mid-dilated position (the so called ``cadaveric position''). Sometimes, some dilation may happen shortly after death because iris muscles relax, but then the pupils become more constricted with the onset of {\it rigor mortis} (post-mortem stiffness) of the constrictor muscles. Also, the pupils can sometimes have different apertures in the left and right eye, if constrictor muscles are influenced by {\it rigor mortis} unevenly \cite{LARPKRAJANG20161}.

No iris reaction to light stimuli can be observed after death, however, the iris may for some time (a few hours) react to selected pharmaceuticals, such as atropine and pilocarpine drops. Iris tissue response to miotic (pupil-constricting) agents was found to be more evident and faster than to mydriatics (pupil-dilating drugs), however, in both cases the reaction is well observed \cite{LARPKRAJANG20161}.

Sometimes, a black line called \emph{tache noire} may appear across the sclera between the eyelids that were left open -- this is due to the drying of tissues exposed to outside conditions. This, however, should not have a direct impact on the appearance of the iris, apart from the damage inflicted to the cornea as a result of eyelids being uncovered.

After longer periods after death (days and weeks), the eyeball will slowly start to lose its firmness due to the progressing loss of intraocular pressure and decomposition of its tissues. This results in a globe that is depressed and flaccid, with severe distortions to its shape \cite{Belsey2016}.

\section{Database of post-mortem iris images }
\label{sec:Database}
The analysis of post-mortem changes and their dynamics carried out for this paper is based upon the database of post-mortem iris images collected at the Warsaw Medical University and distributed by Warsaw University of Technology \cite{WarsawColdIris1}. The corpus comprises iris images collected in mortuary conditions from 17 subjects, \ie 34 distinct irises are represented in the dataset. The longest available observation horizon is 814 hours after death (almost 34 days). Near infrared images using IriShield M2120U sensor (640$\times$480 size) and visible light color images (16MP size) using a general-purpose camera are available in this dataset. NIR and visible light samples were taken almost simultaneously, allowing the analyses in different light spectra for each subject.

\section{Post-mortem iris decomposition}
\label{sec:Analysis}

\subsection{Global Features}
\label{sec:Global}

In this section, we analyze dynamics and characteristics of post-mortem changes in the iris and its surroundings from a coarse perspective, and qualitatively assessing eye degradation. Figures included below represent near-infrared \textbf{(always on the left)} and visible light \textbf{(always on the right)} samples of the right eye of subject 17 from the database \cite{WarsawColdIris1}. This is the only subject with such a long observation horizon, reaching 814 hours of post-mortem interval. Each figure caption provides a brief commentary on the iris degradation in each moment. All pictures are best viewed in color.

\begin{figure}[!h]
	\centering
	\includegraphics[width=0.401\textwidth]{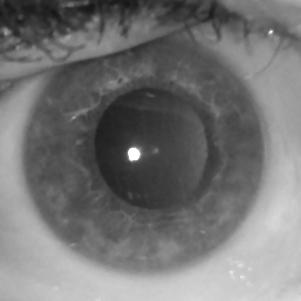}\hskip1mm
	\includegraphics[width=0.401\textwidth]{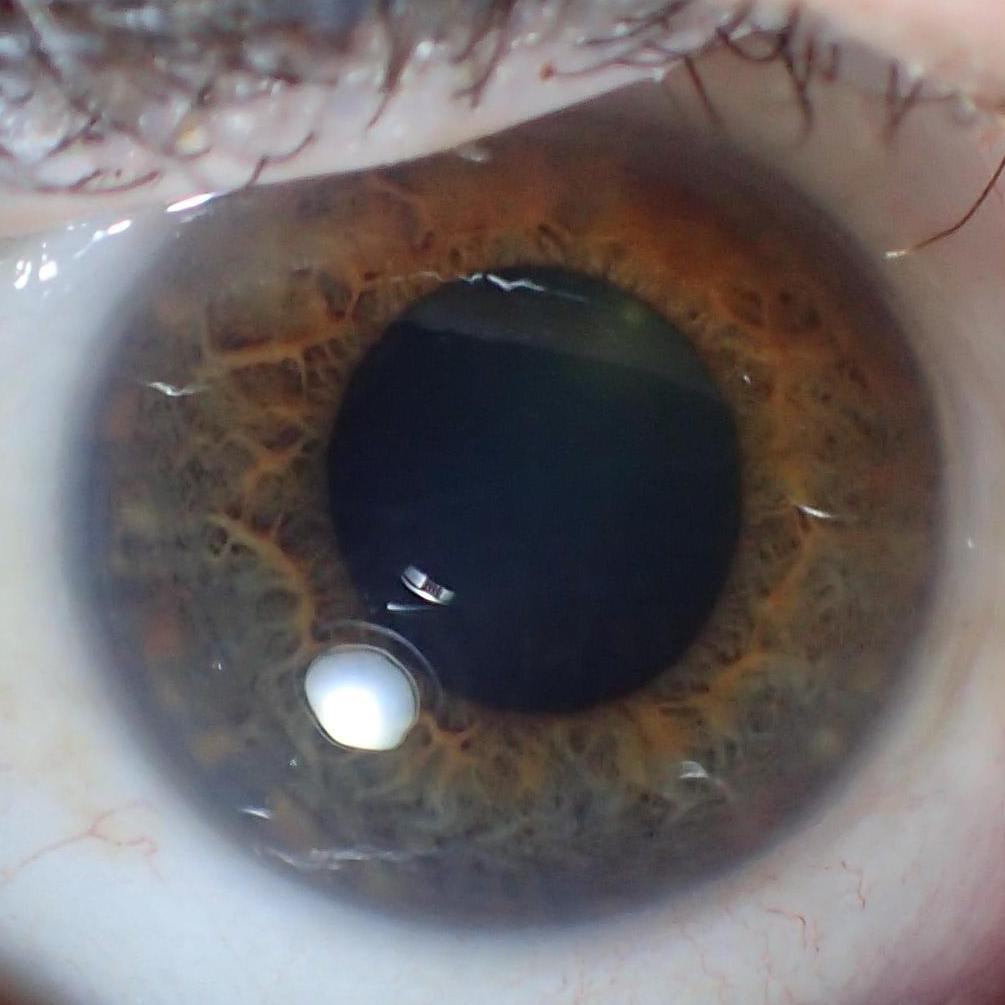}\\
	\caption{{\bf5 hours post-mortem:} no apparent changes to the iris texture in either NIR or RGB.}
	\label{fig:global-s1}
\end{figure}

\begin{figure}[!h]
	\centering
	\includegraphics[width=0.401\textwidth]{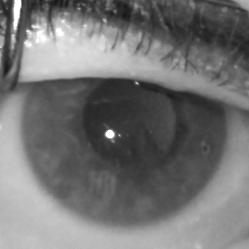}\hskip1mm
	\includegraphics[width=0.401\textwidth]{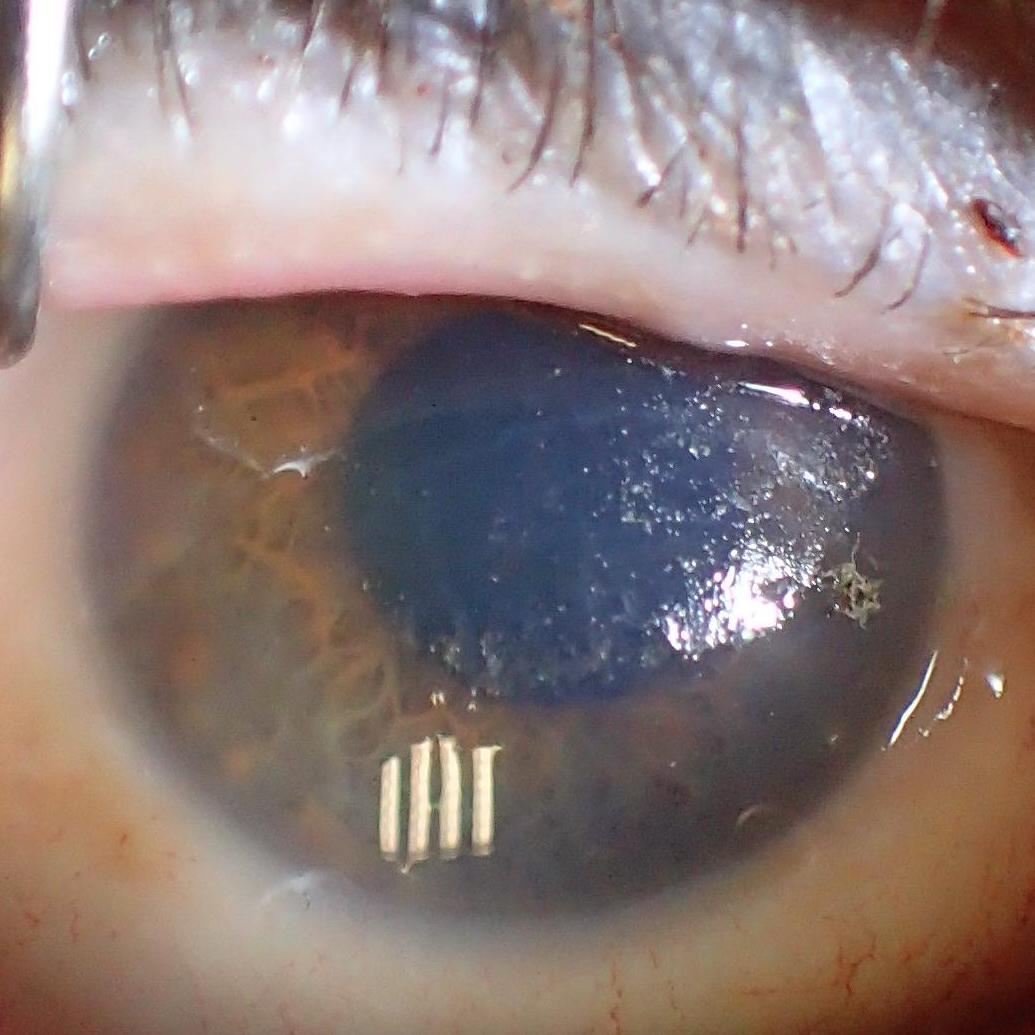}\\
	\caption{{\bf 23 hours post-mortem:} corneal opacification and drying appears, visible in RGB. NIR seems to be more robust to these changes, but the iris features are starting to blur.}
	\label{fig:global-s2}
\end{figure}

\begin{figure}[!h]
	\centering
	\includegraphics[width=0.401\textwidth]{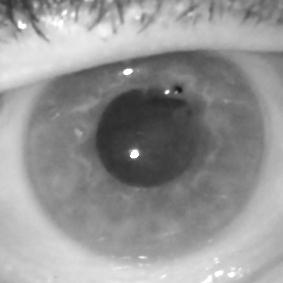}\hskip1mm
	\includegraphics[width=0.401\textwidth]{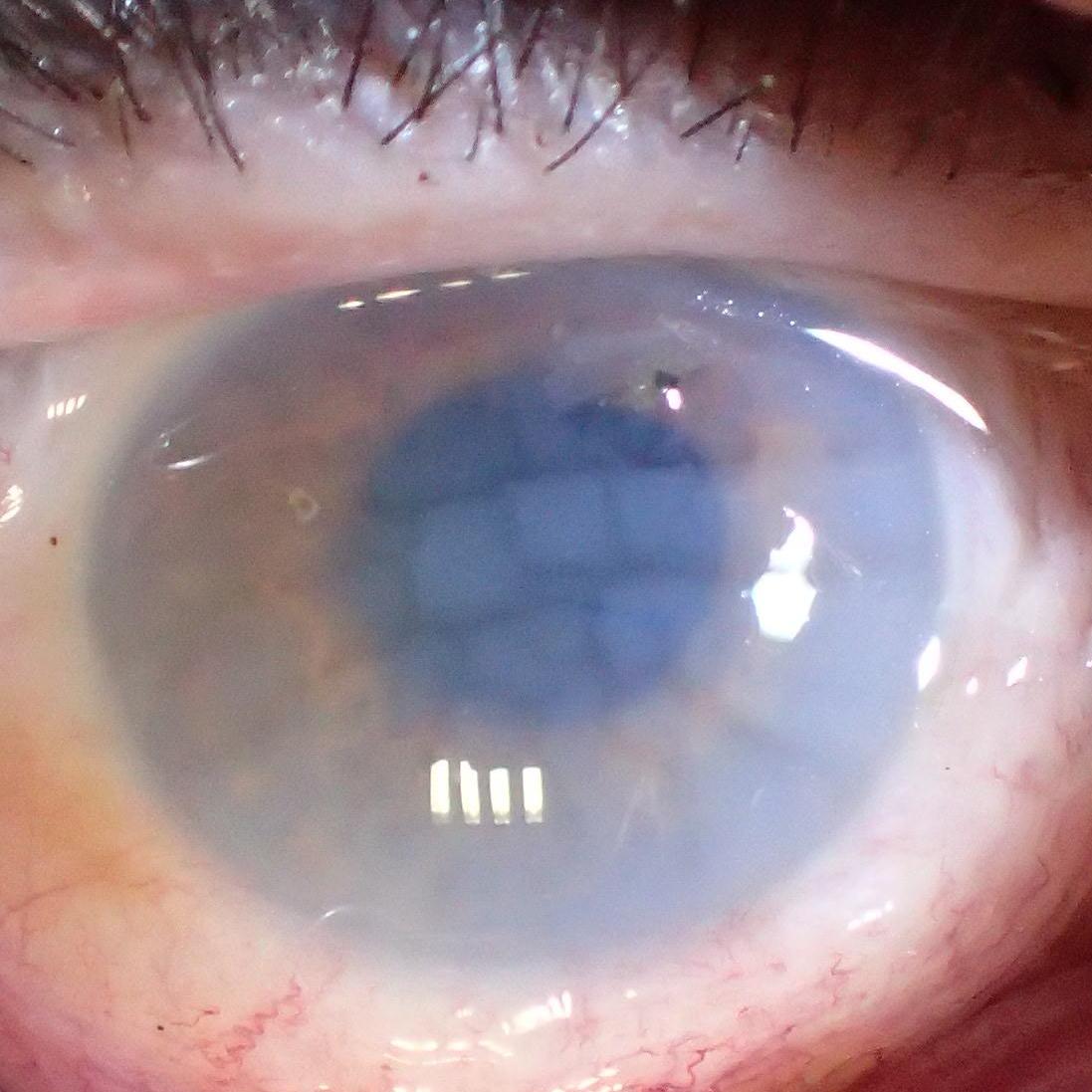}\\
	\caption{{\bf 95 hours post-mortem:} NIR image looks unaffected by post-mortem changes, but they are getting more pronounced in the RGB, where corneal opacification is progressing, and wrinkling begins manifesting itself. Some features are still distinguishable, albeit most appear blurry.}
	\label{fig:global-s3}
\end{figure}

\begin{figure}[!h]
	\centering
	\includegraphics[width=0.391\textwidth]{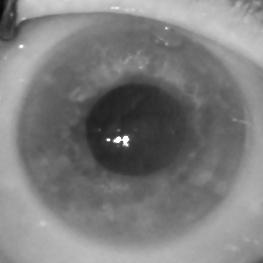}\hskip1mm
	\includegraphics[width=0.391\textwidth]{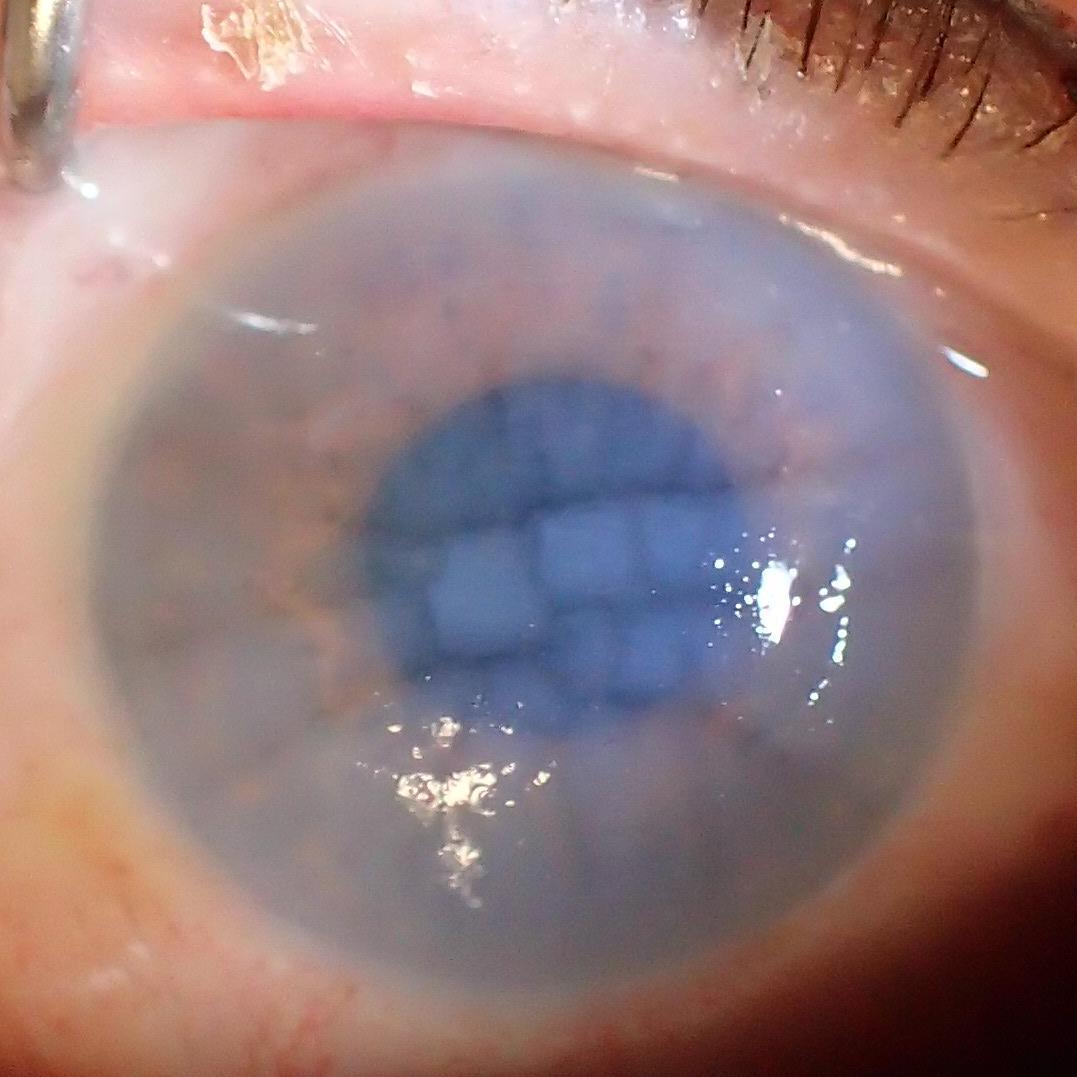}\\
	\caption{{\bf 154 hours post-mortem:} NIR still shows iris features, whereas opacification and wrinkling of the cornea is visible in RGB. Iris features possible to be used in identification are still visible in both images.}
	\label{fig:global-s4}
\end{figure}

\begin{figure}[!h]
	\centering
	\includegraphics[width=0.391\textwidth]{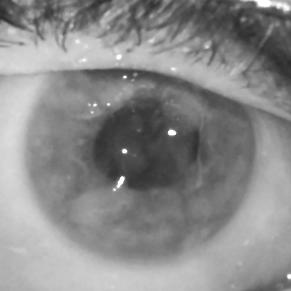}\hskip1mm
	\includegraphics[width=0.391\textwidth]{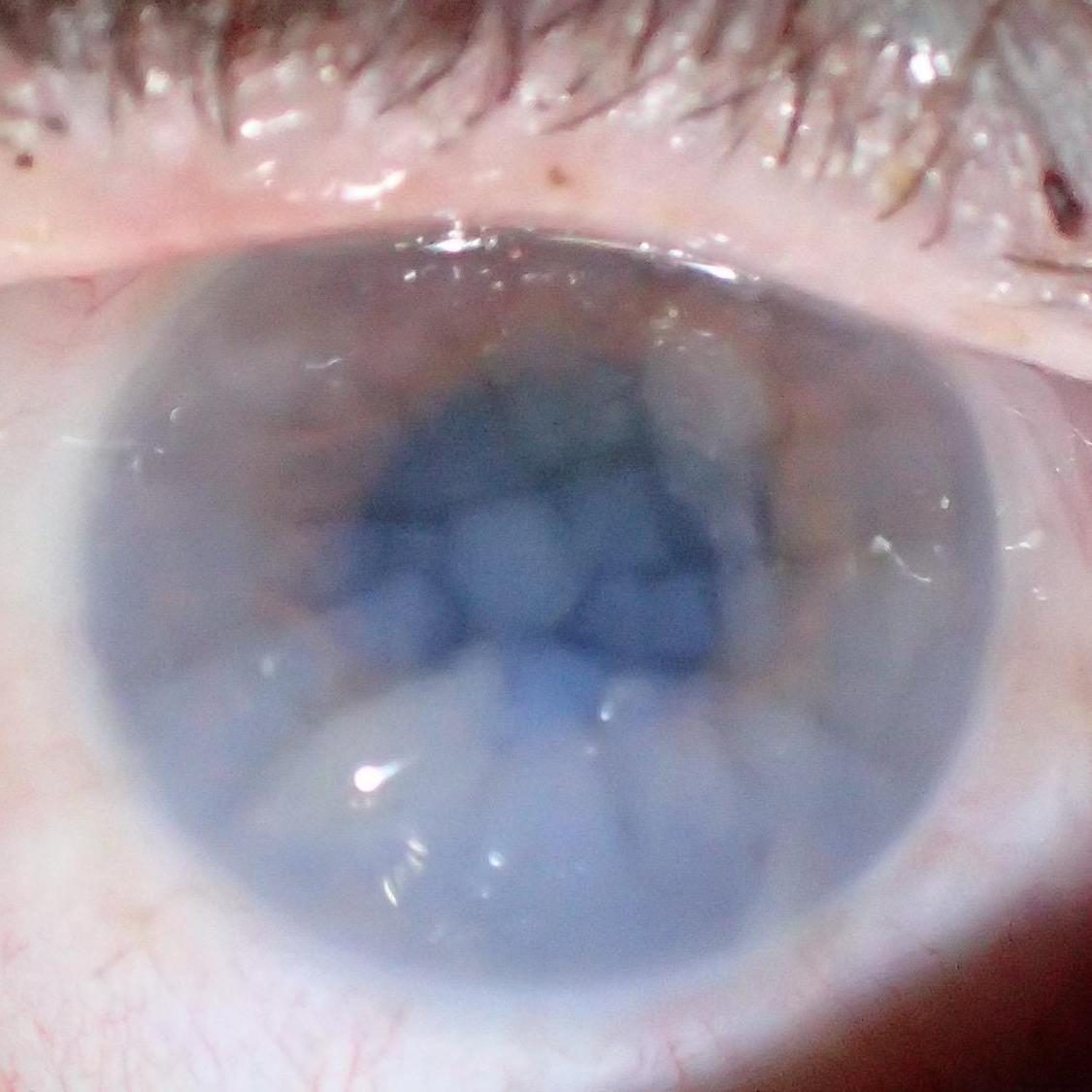}\\
	\caption{{\bf 215 hours post-mortem:} the opacification can be seen in NIR for the first time, causing additional light reflections. The iris texture is still observable. RGB image is significantly affected by opacification and wrinkling, blurring most of the iris texture.}
	\label{fig:global-s5}
\end{figure}

\begin{figure}[!h]
	\centering
	\includegraphics[width=0.391\textwidth]{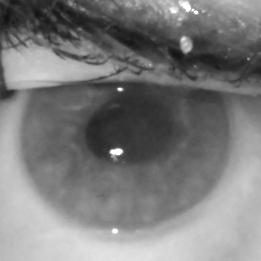}\hskip1mm
	\includegraphics[width=0.391\textwidth]{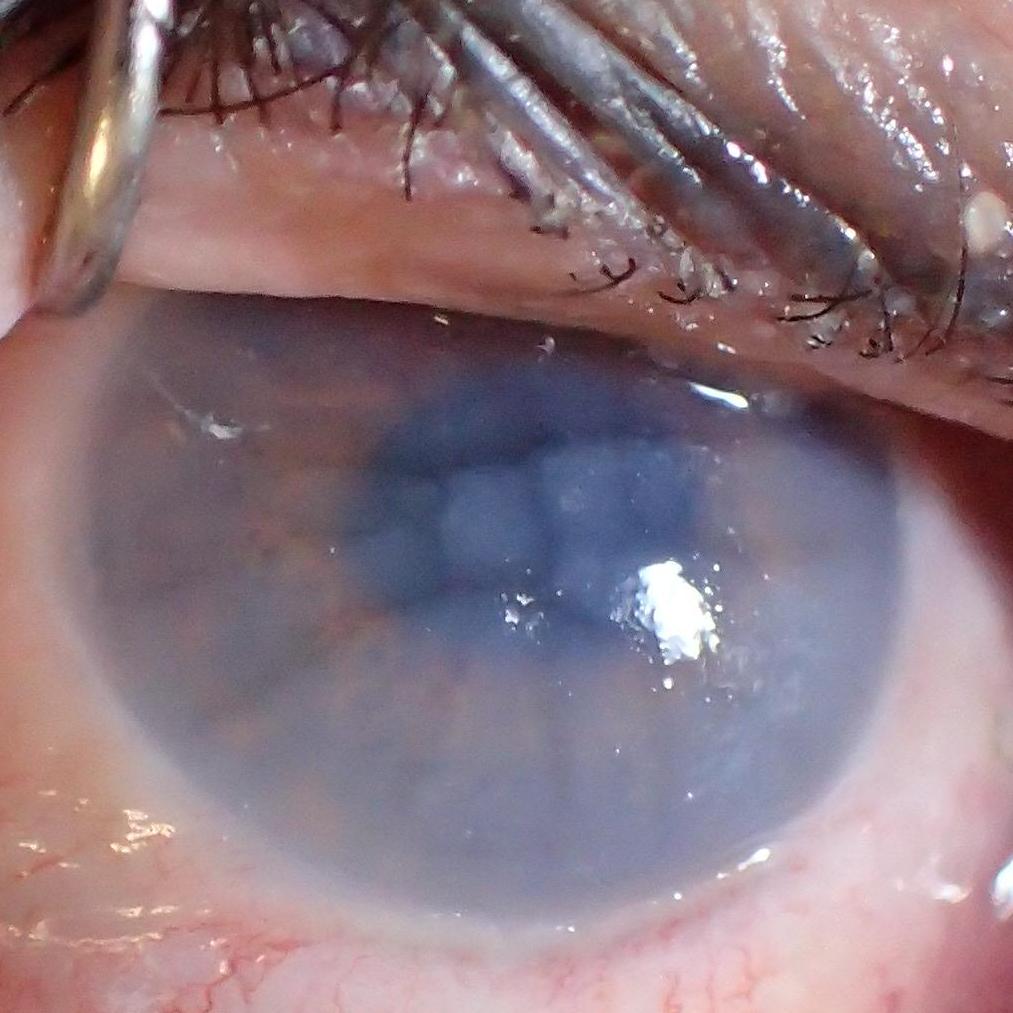}\\
	\caption{{\bf 263 hours post-mortem:} both images look clearer than in the previous session, most likely thanks to better image exposure and focus. NIR offers good visibility of features, RGB only in selected (leftmost) regions.}
	\label{fig:global-s6}
\end{figure}

\begin{figure}[!h]
	\centering
	\includegraphics[width=0.381\textwidth]{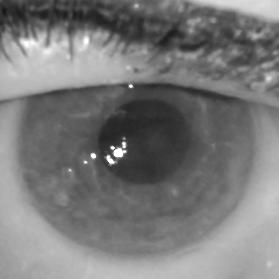}\hskip1mm
	\includegraphics[width=0.381\textwidth]{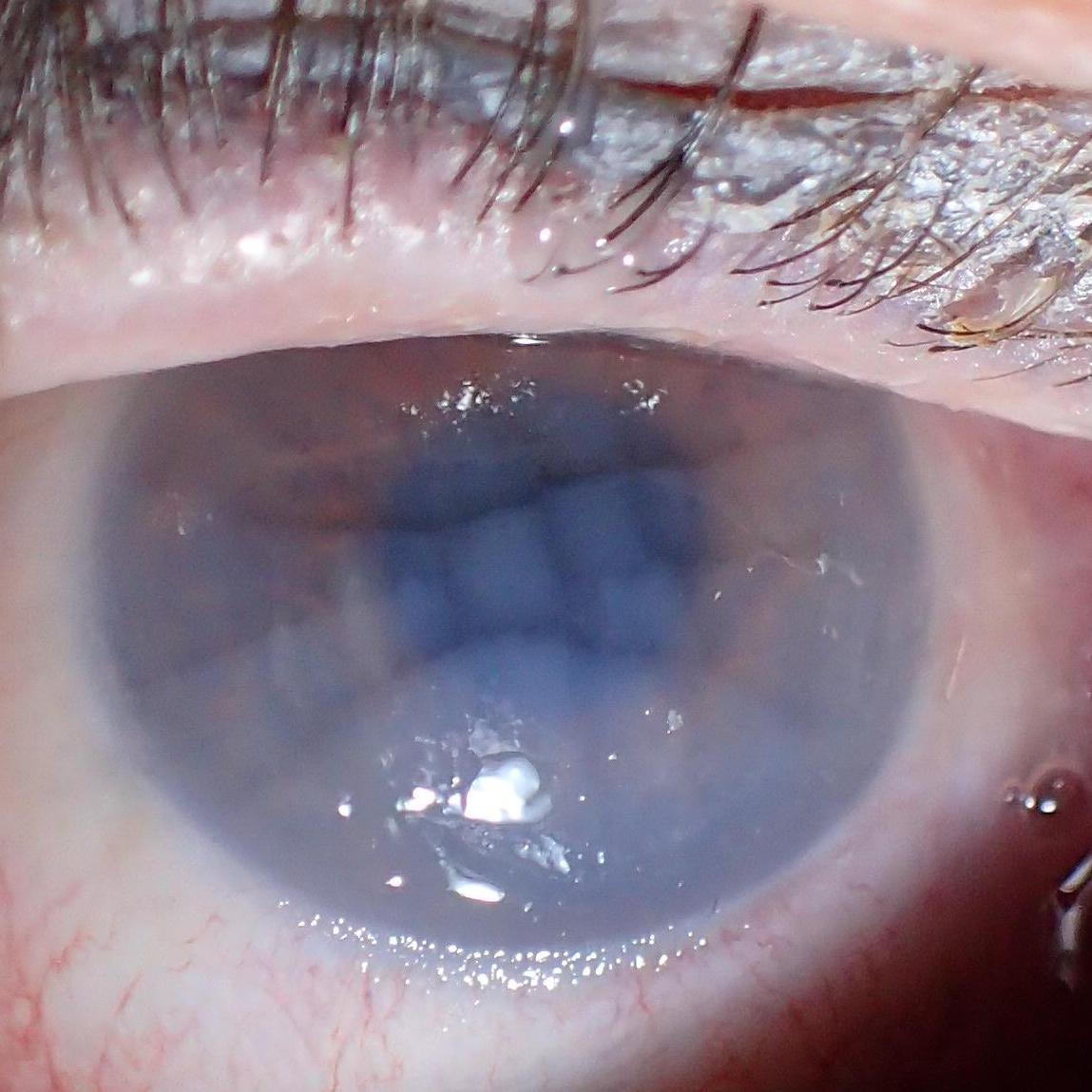}\\
	\caption{{\bf 359 hours post-mortem:} similar to previous session, but with more light reflections and corneal wrinkling, which is now also seen in NIR. Iris features still observable in NIR, as well as in some portions of the RGB image.}
	\label{fig:global-s7}
\end{figure}

\begin{figure}[!h]
	\centering
	\includegraphics[width=0.381\textwidth]{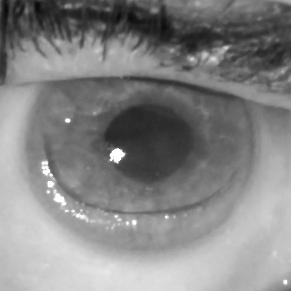}\hskip1mm
	\includegraphics[width=0.381\textwidth]{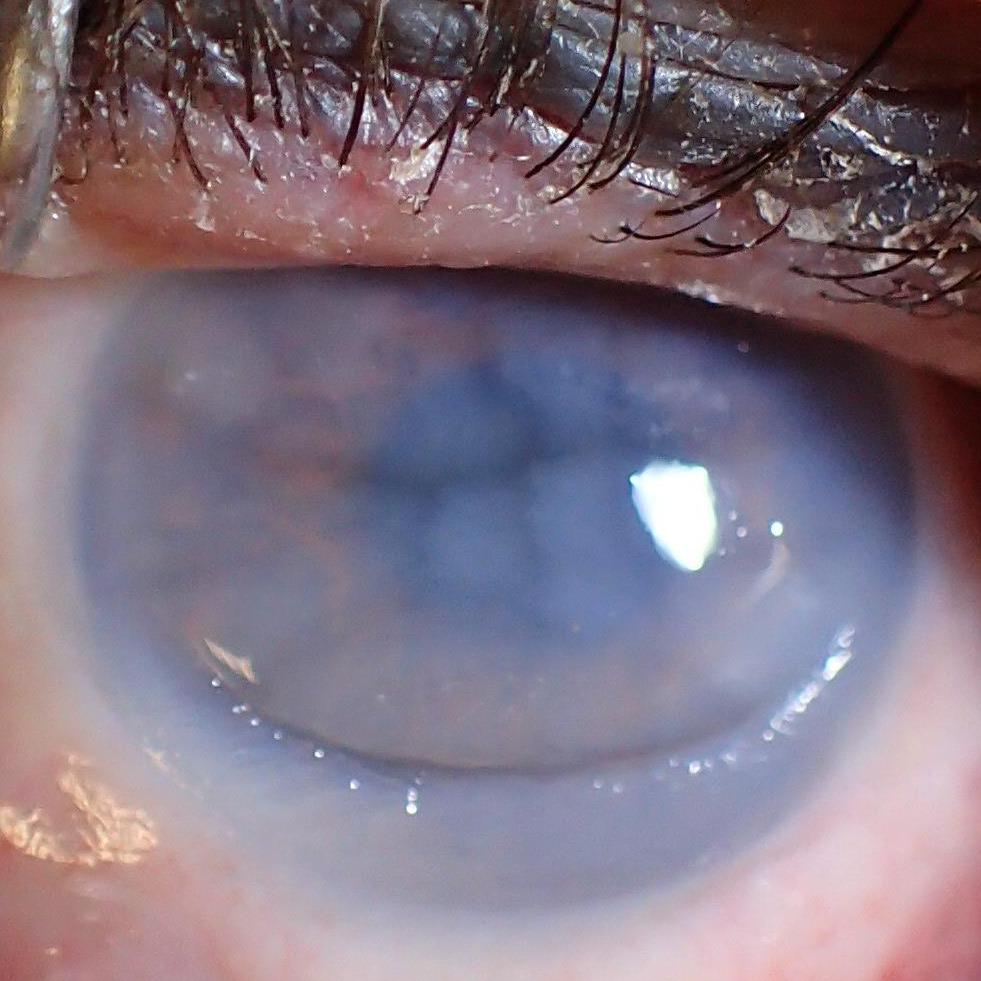}\\
	\caption{{\bf 407 hours post-mortem:} the eyeball collapse is observed in this session with a big wrinkle in the bottom part, eliminating features in this region. The upper part still visible in NIR. Thanks to good focusing, some iris features are still visible in the RGB image.} 
	\label{fig:global-s8}
\end{figure}

\begin{figure}[!h]
	\centering
	\includegraphics[width=0.381\textwidth]{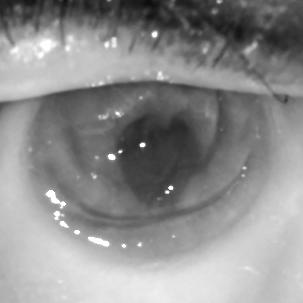}\hskip1mm
	\includegraphics[width=0.381\textwidth]{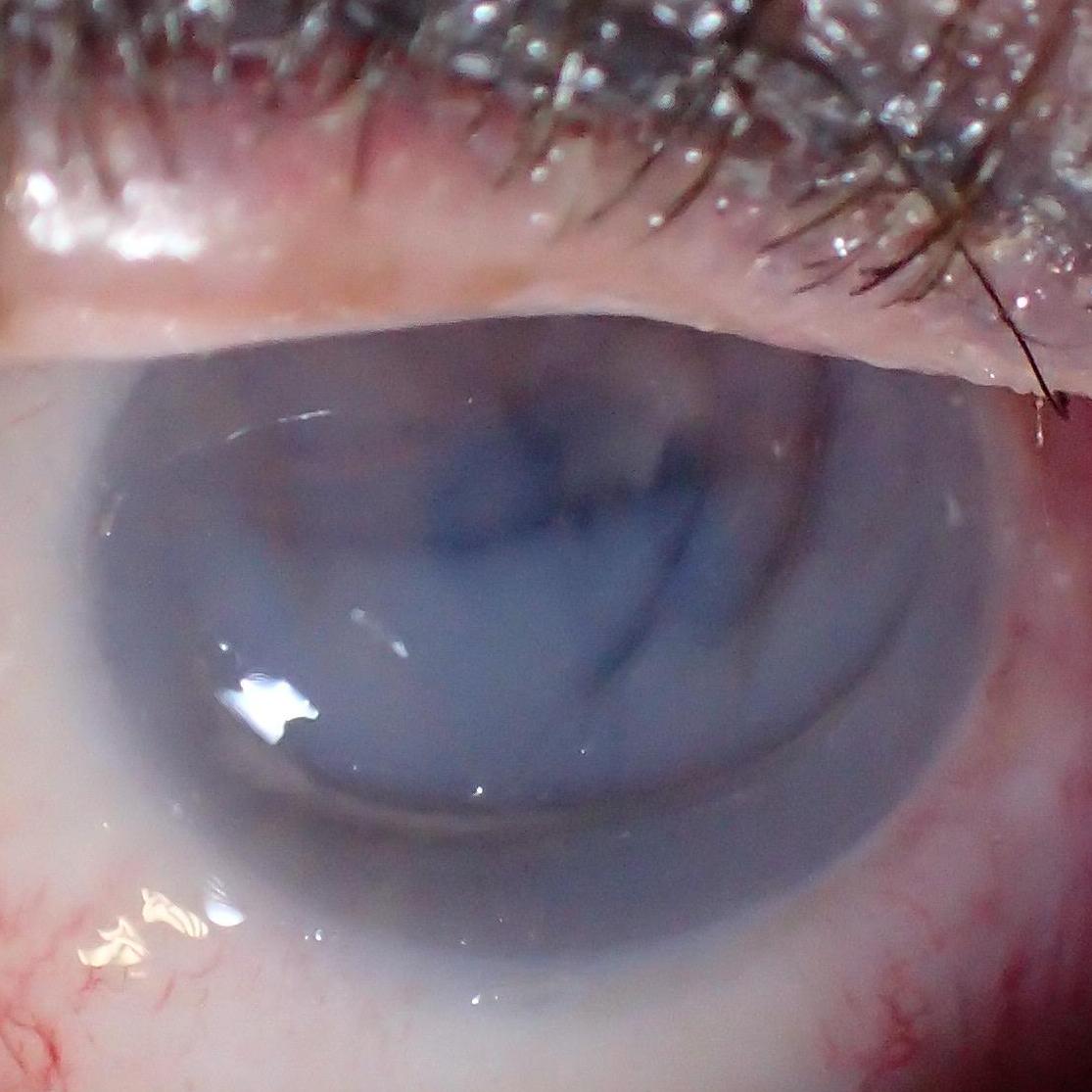}\\
	\caption{{\bf 503 hours post-mortem:} the eyeball collapse visible in both images, in combination with corneal haze and wrinkling, affects almost all iris features. Some very limited portions of the iris can perhaps be still retrieved in NIR.}
	\label{fig:global-s9}
\end{figure}

\begin{figure}[!h]
	\centering
	\includegraphics[width=0.401\textwidth]{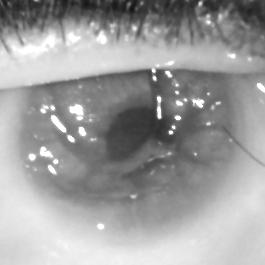}\hskip1mm
	\includegraphics[width=0.401\textwidth]{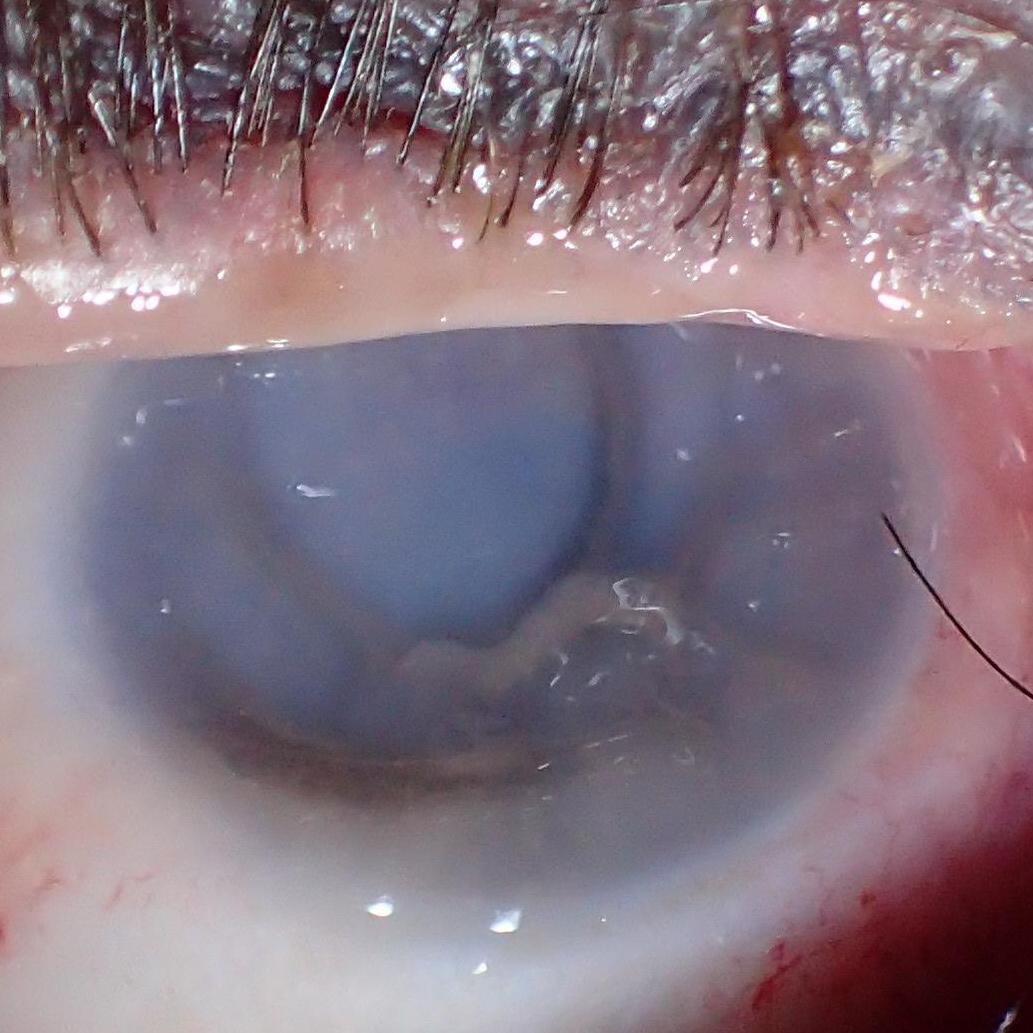}\\
	\caption{{\bf 574 hours post-mortem:} the RGB image depicts a total lack of iris features, now hidden beneath the haze of a desiccated and wrinkled cornea. The NIR image shows a small portion of the iris (upper part directly below the eyelid) that perhaps still has \emph{some} features useful in identification.}
	\label{fig:global-s10}
\end{figure}

\begin{figure}[!h]
	\centering
	\includegraphics[width=0.401\textwidth]{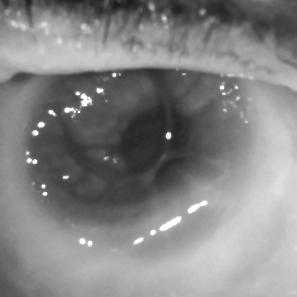}\hskip1mm
	\includegraphics[width=0.401\textwidth]{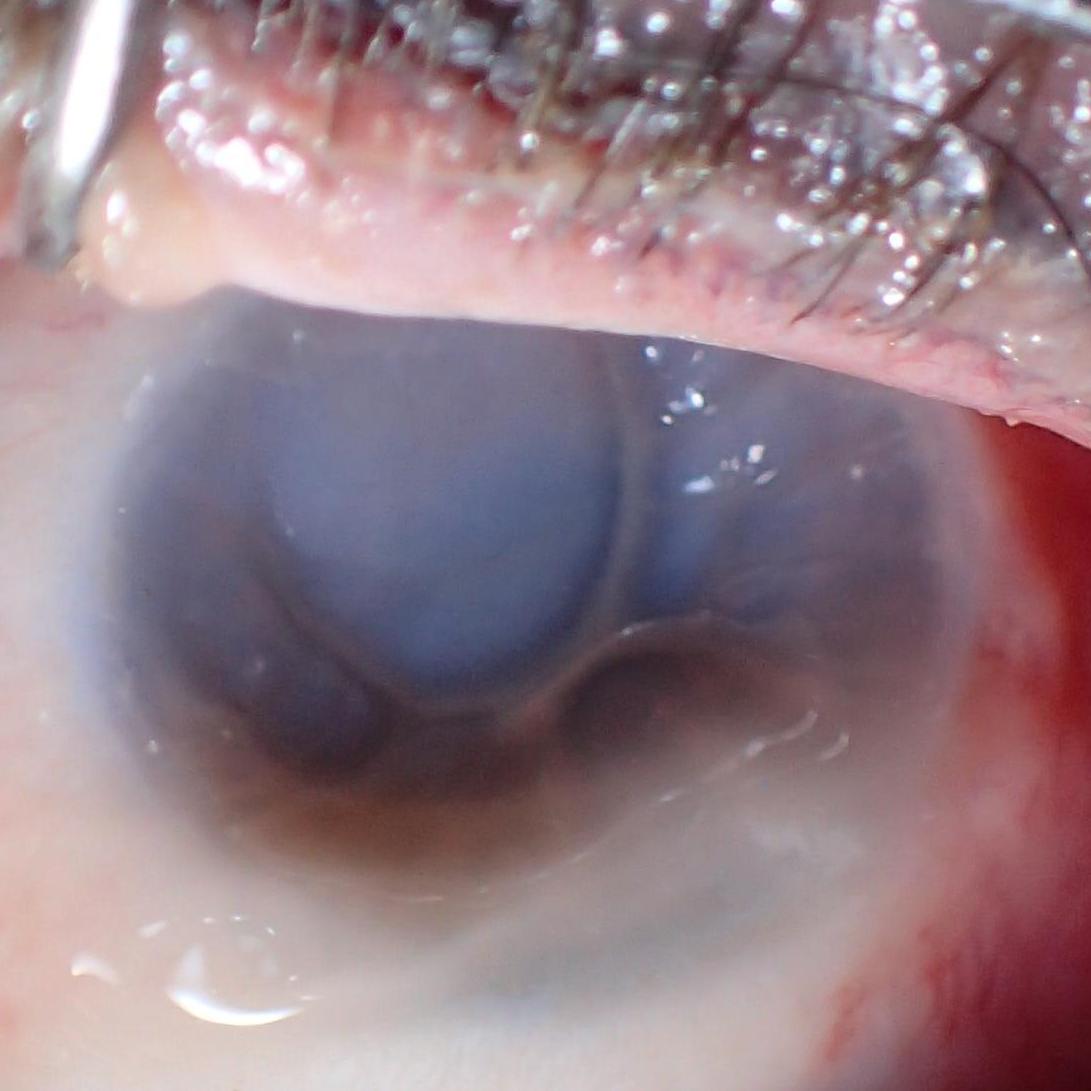}\\
	\caption{{\bf 622 hours post-mortem:} The RGB image now seems useless. The small portion of iris features seemingly preserved in NIR are getting more blurry. Opacification and wrinkling progresses, with eye drying out.}
	\label{fig:global-s11}
\end{figure}

\begin{figure}[!h]
	\centering
	\includegraphics[width=0.401\textwidth]{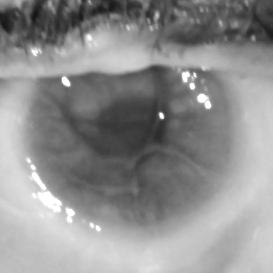}\hskip1mm
	\includegraphics[width=0.401\textwidth]{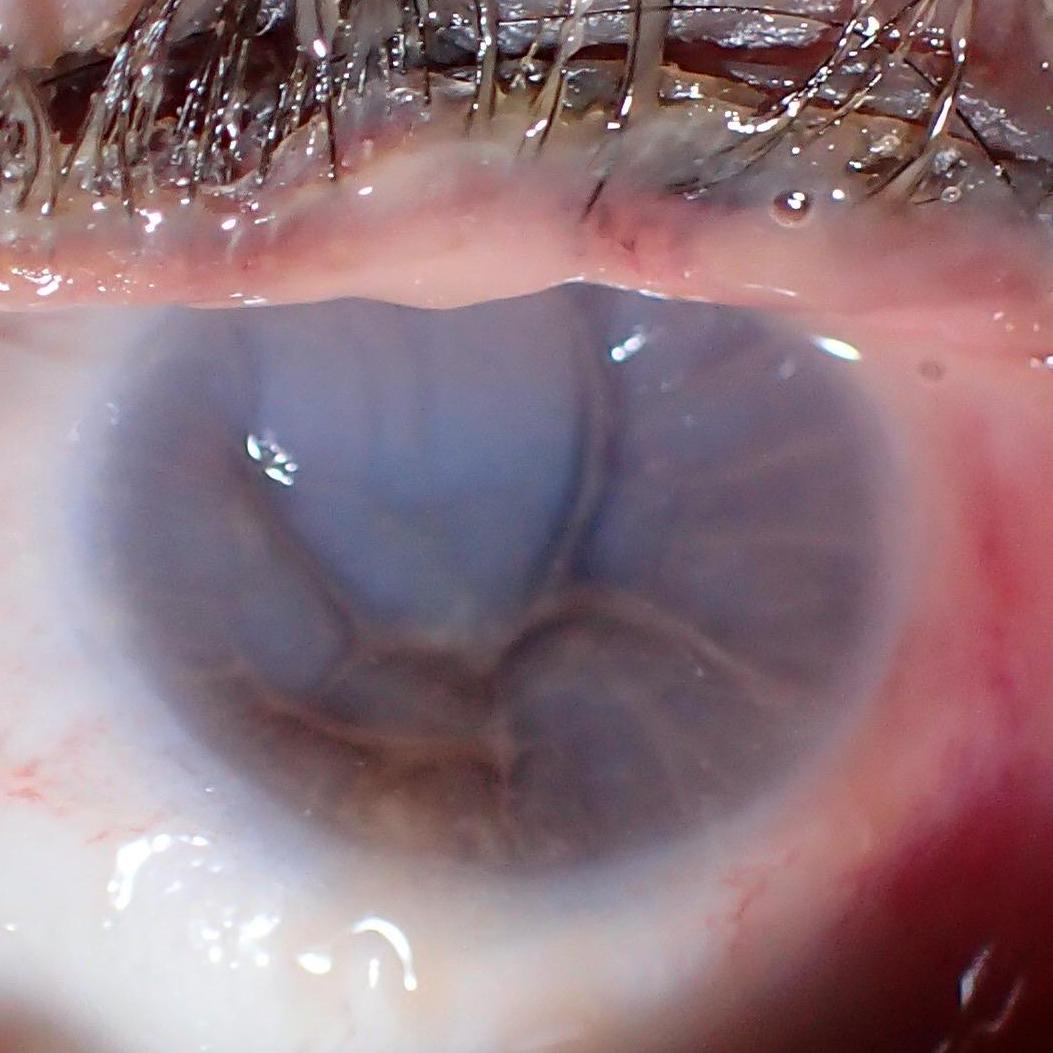}\\
	\caption{{\bf 742 hours post-mortem:} no distinctive iris features in both images can be visible.}
	\label{fig:global-s12}
\end{figure}

\begin{figure}[!h]
	\centering
	\includegraphics[width=0.401\textwidth]{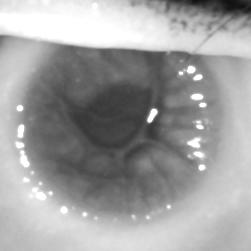}\hskip1mm
	\includegraphics[width=0.401\textwidth]{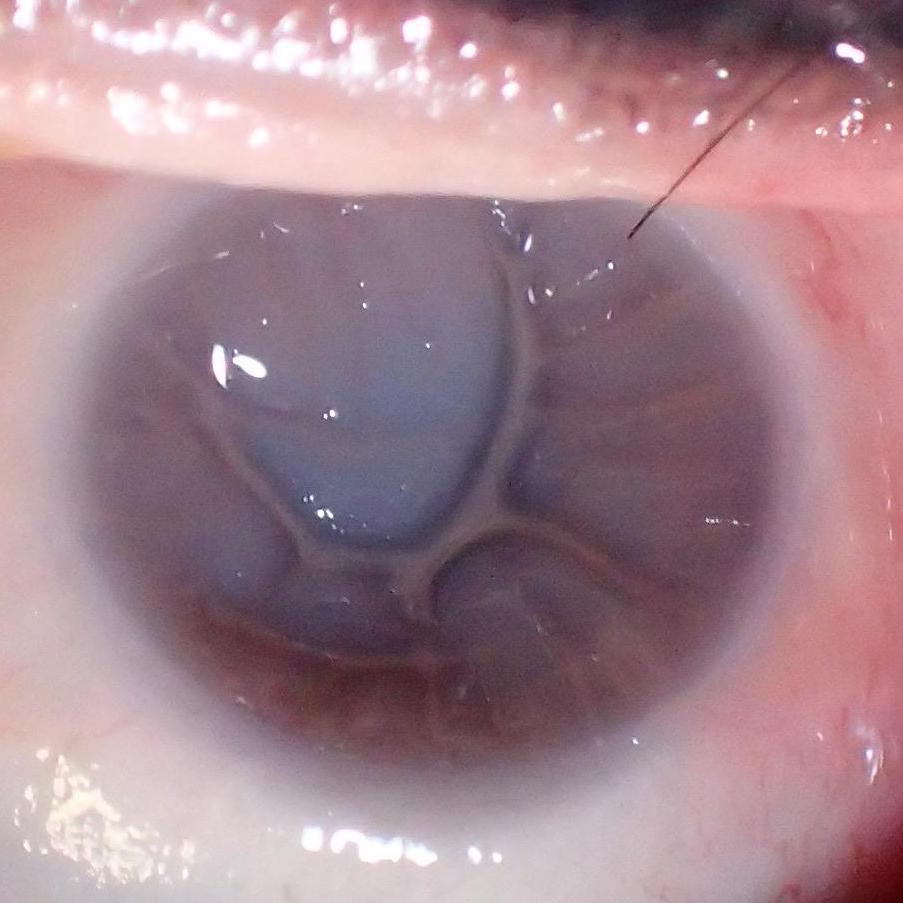}\\
	\caption{{\bf 814 hours post-mortem:} no distinctive iris features in both images can be visible.}
	\label{fig:global-s13}
\end{figure}

This analysis delivers a few important observations, which may be useful and interesting from the forensic viewpoint:
\begin{enumerate}
    \item \textbf{Image illumination.} Near-infrared illumination has been found to be able to better penetrate the corneal opacification, or 'haze', thus offering much better visibility of the iris texture, even when considering long observation horizons. For the samples analyzed above, the iris image acquired in NIR light may in some regions look similar to what we would expect from a living iris, even as long as 407 hours after the subject's demise. A visible light images already depict an intensive haze and wrinkling, which washes out most of the iris texture.
    \item \textbf{Image resolution.} Although the NIR images seem genuinely better for this task, they lack the high resolution of the RGB images, which causes the iris features to become blurry faster than in RGB images. An ideal solution would be to employ a high-resolution near-infrared camera for imaging post-mortem samples, offering a forensic expert as much data as possibly can be obtained.
    \item \textbf{Borderline session.} For this subject, a borderline session can be subjectively identified as the one still offering good iris features visibility, here it is 407 hours, or 17 days, post-mortem. After this time horizon is crossed, this particular iris will likely be no longer useful for the purpose of recognition due to severe tissue degradation, resulting in almost total absence of good quality iris texture. 
\end{enumerate}

\begin{figure*}[h!]
	\centering 
	{\bf Feature annotation in NIR image:}\\\vskip1mm
	    \includegraphics[width=0.12\textwidth]{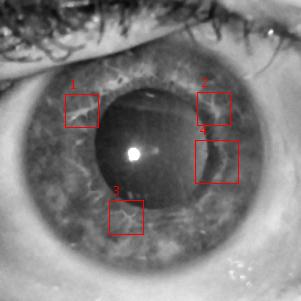}
	\includegraphics[width=0.12\textwidth]{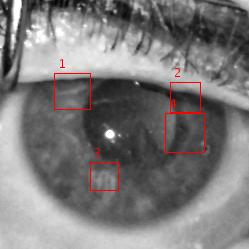}
	\includegraphics[width=0.12\textwidth]{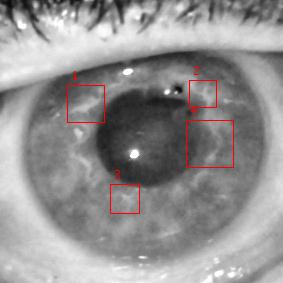}
	\includegraphics[width=0.12\textwidth]{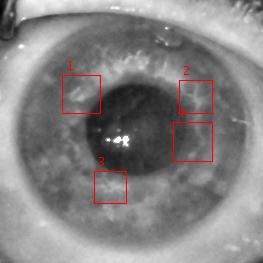}
	\includegraphics[width=0.12\textwidth]{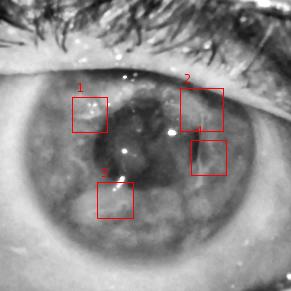}
	\includegraphics[width=0.12\textwidth]{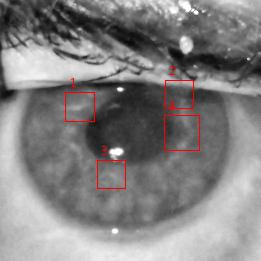}
	\includegraphics[width=0.12\textwidth]{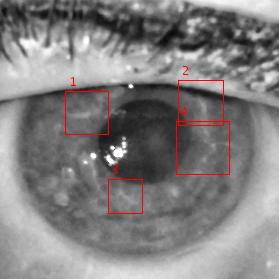}
	\includegraphics[width=0.12\textwidth]{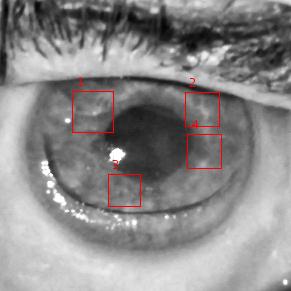}\\
	5 hours \hskip8mm 23 hours \hskip8mm 95 hours \hskip8mm 154 hours \hskip8mm 215 hours \hskip8mm 263 hours \hskip8mm 359 hours \hskip8mm 407 hours \\\vskip2mm
	{\bf Individual feature close-ups:}\\
	\hskip-107mm Feature 1:\\\vskip1mm
		\includegraphics[width=0.12\textwidth]{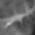}
	\includegraphics[width=0.12\textwidth]{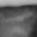}
	\includegraphics[width=0.12\textwidth]{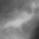}
	\includegraphics[width=0.12\textwidth]{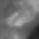}
	\includegraphics[width=0.12\textwidth]{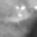}
	\includegraphics[width=0.12\textwidth]{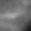}
	\includegraphics[width=0.12\textwidth]{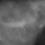}
	\includegraphics[width=0.12\textwidth]{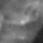}\\
	\hskip-107mm Feature 2:\\\vskip1mm
	    \includegraphics[width=0.12\textwidth]{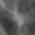}
	\includegraphics[width=0.12\textwidth]{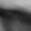}
	\includegraphics[width=0.12\textwidth]{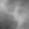}
	\includegraphics[width=0.12\textwidth]{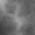}
	\includegraphics[width=0.12\textwidth]{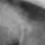}
	\includegraphics[width=0.12\textwidth]{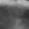}
	\includegraphics[width=0.12\textwidth]{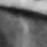}
	\includegraphics[width=0.12\textwidth]{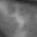}\\
	\hskip-107mm Feature 3:\\\vskip1mm
	    \includegraphics[width=0.12\textwidth]{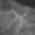}
	\includegraphics[width=0.12\textwidth]{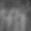}
	\includegraphics[width=0.12\textwidth]{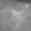}
	\includegraphics[width=0.12\textwidth]{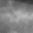}
	\includegraphics[width=0.12\textwidth]{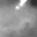}
	\includegraphics[width=0.12\textwidth]{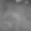}
	\includegraphics[width=0.12\textwidth]{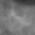}
	\includegraphics[width=0.12\textwidth]{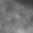}\\
	\hskip-107mm Feature 4:\\\vskip1mm
	    \includegraphics[width=0.12\textwidth]{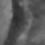}
	\includegraphics[width=0.12\textwidth]{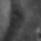}
	\includegraphics[width=0.12\textwidth]{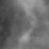}
	\includegraphics[width=0.12\textwidth]{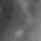}
	\includegraphics[width=0.12\textwidth]{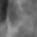}
	\includegraphics[width=0.12\textwidth]{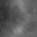}
	\includegraphics[width=0.12\textwidth]{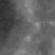}
	\includegraphics[width=0.12\textwidth]{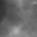}
	\caption{Close-ups of four features manually annotated throughout 8 acquisition sessions until the borderline one.}
	\label{fig:closeup-NIR}
\end{figure*}

\subsection{Close-up Analysis}
\label{sec:CloseUp}

For the same eye used in the `global' analysis, a close-up inspection is then performed, to assess metamorphoses and degradation on the iris feature level, \ie how the individual iris details such as collagen fibers and crypts withstand the elapsing time post-mortem. To proceed with this, we have manually chosen and annotated four corresponding iris features in each image over the course of acquisition sessions up to the borderline session, for both illumination types.

Four features annotated in the NIR images are shown in Fig. \ref{fig:closeup-NIR}. For easier perception of the iris features, especially in the later sessions, contrast-limited adaptive histogram equalization (CLAHE) preprocessing is applied to the NIR image. Features 2 and 4 seem to be resilient to post-mortem changes, as their shapes can still be discerned even in the eighth and seventh session (407 hours after death). On the other hand, Features 1 and 3 are getting blurry rather fast, degrading quickly. Also, the image collected during the second session (23 hours post-mortem) seems to have especially bad quality, which is most likely attributed to bad focus, rather than iris degradation, as the iris looks much cleaner in the following sessions 3 and 4. This shows the importance of getting the best possible image when aiming for post-mortem iris recognition, as post-mortem processes combined with degraded image quality can severely affect the appearance of iris texture.

\begin{figure*}[t]
	\includegraphics[width=0.16\textwidth]{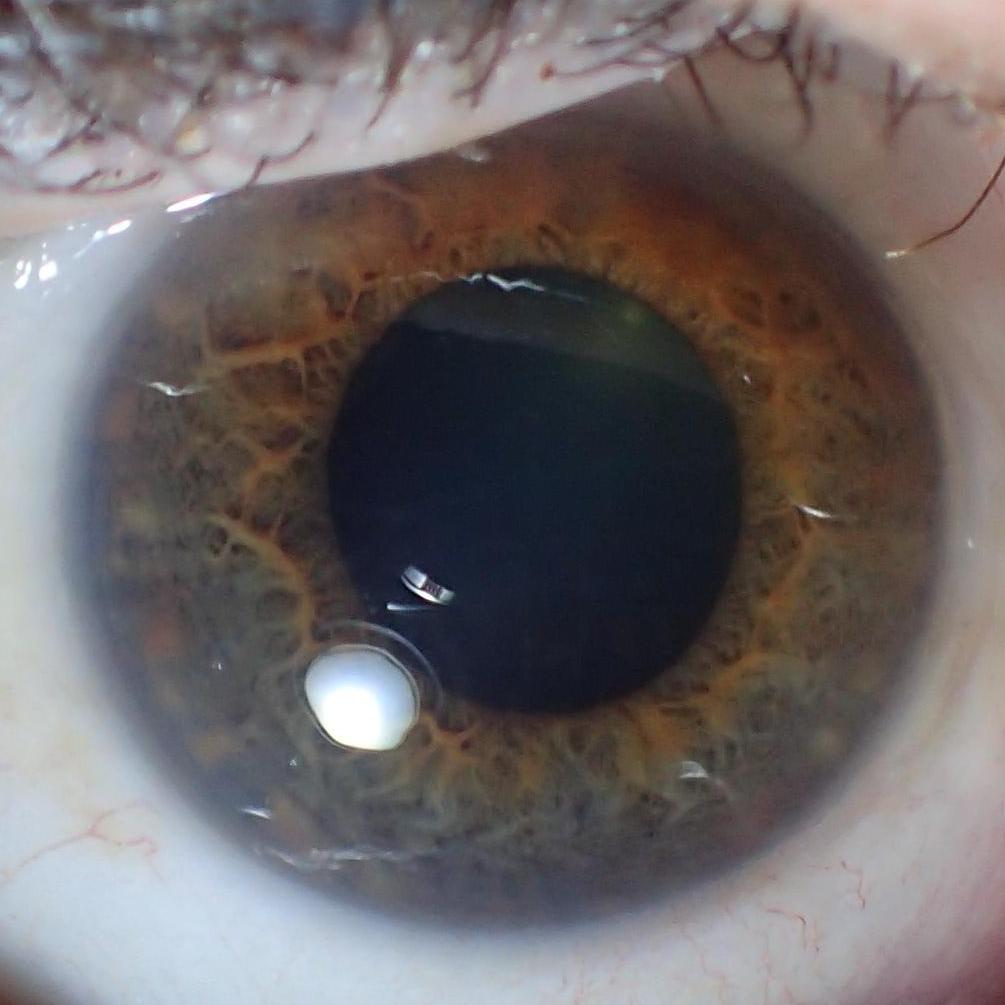}\hfill
	\includegraphics[width=0.16\textwidth]{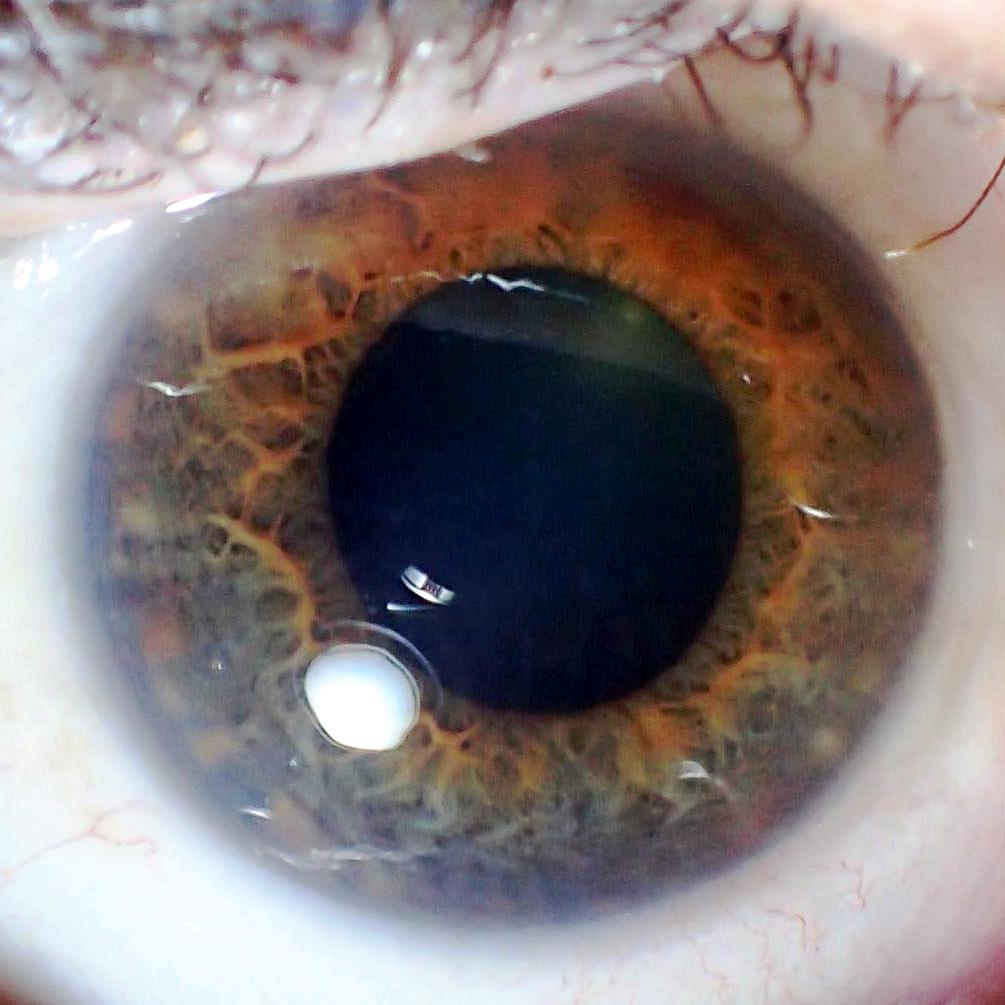}
	\includegraphics[width=0.16\textwidth]{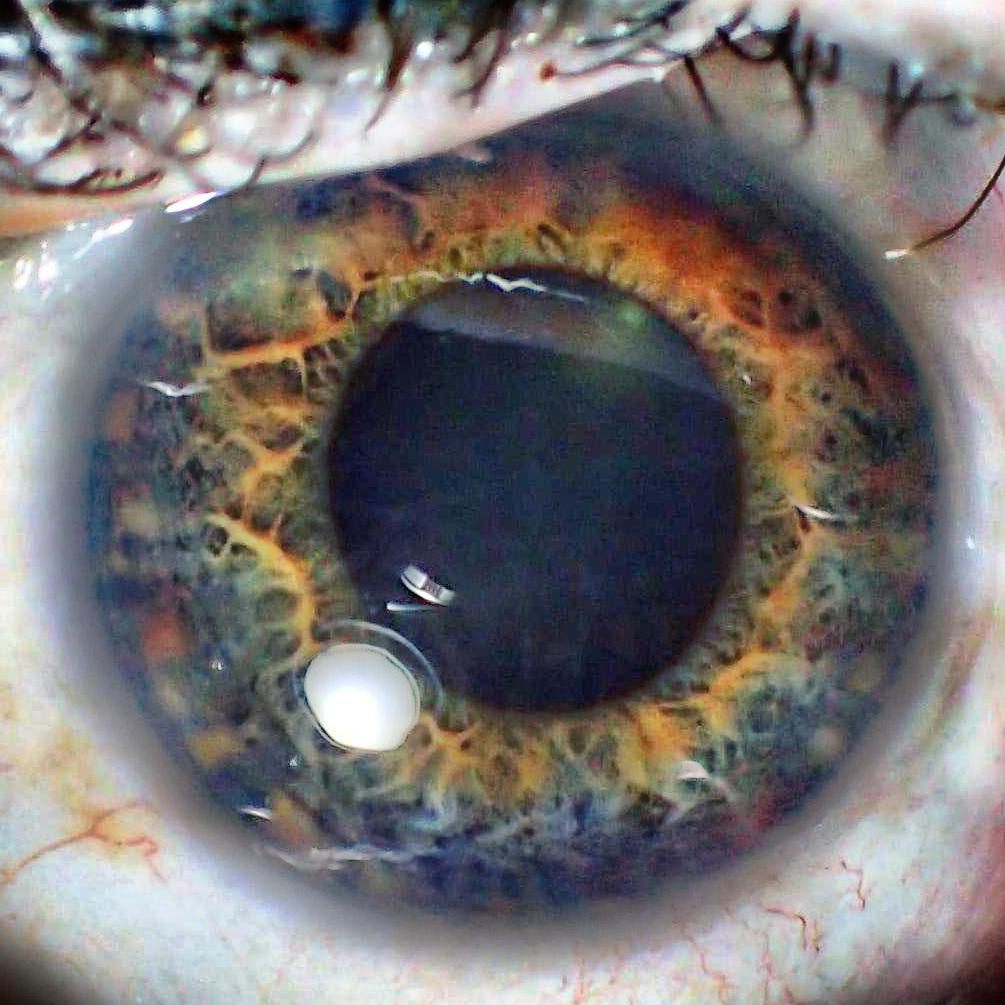}
	\includegraphics[width=0.16\textwidth]{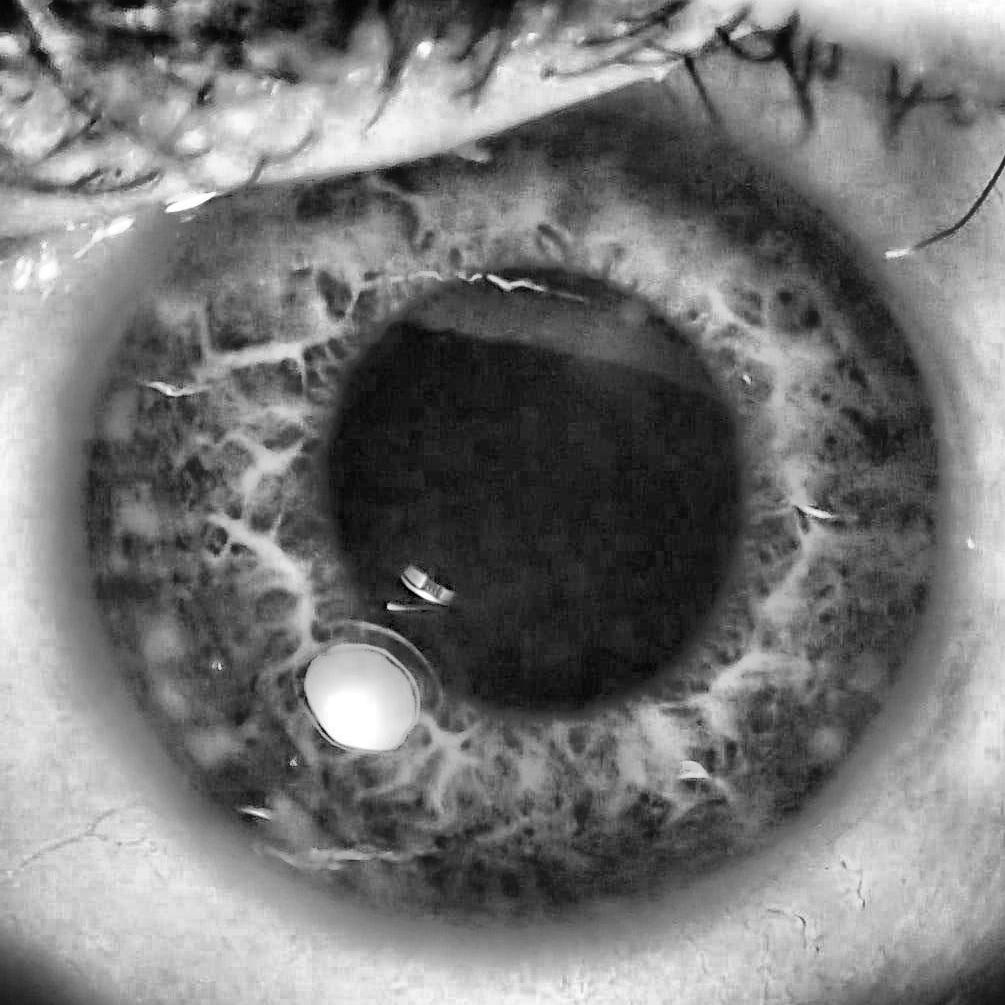}
	\includegraphics[width=0.16\textwidth]{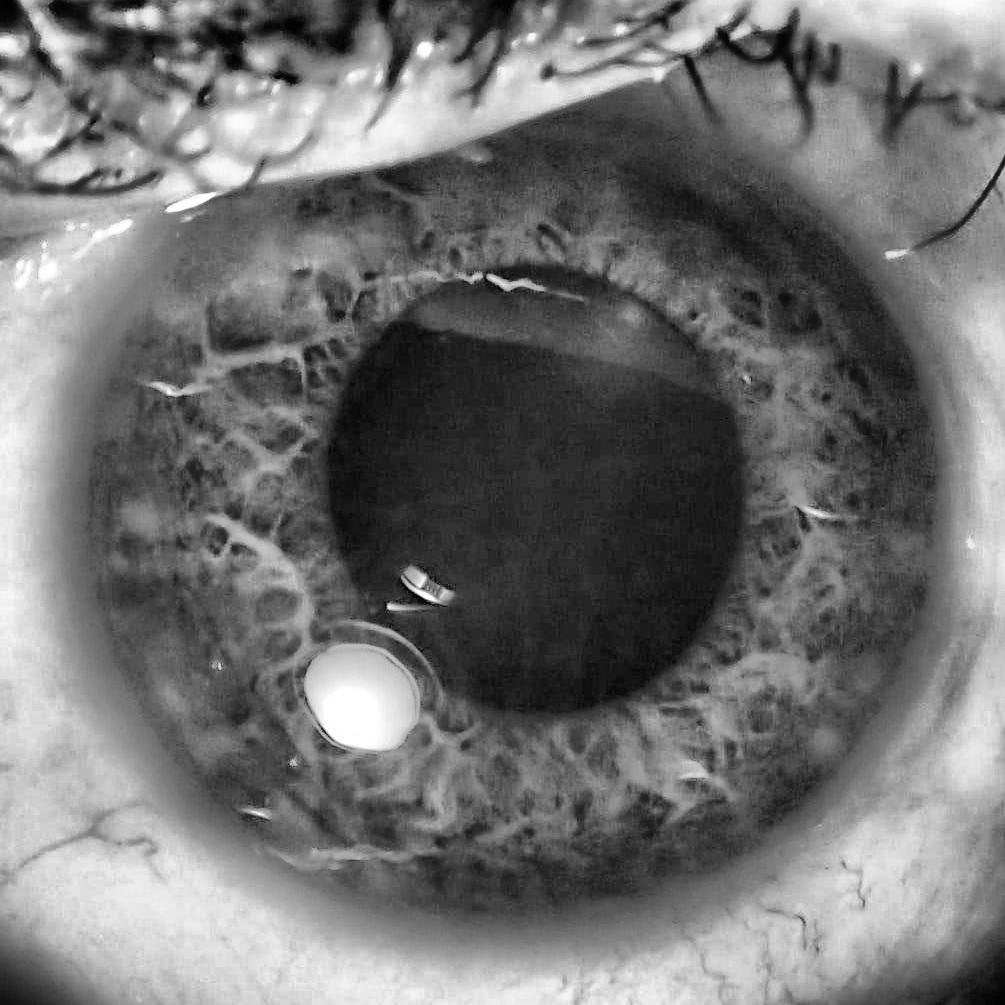}
	\includegraphics[width=0.16\textwidth]{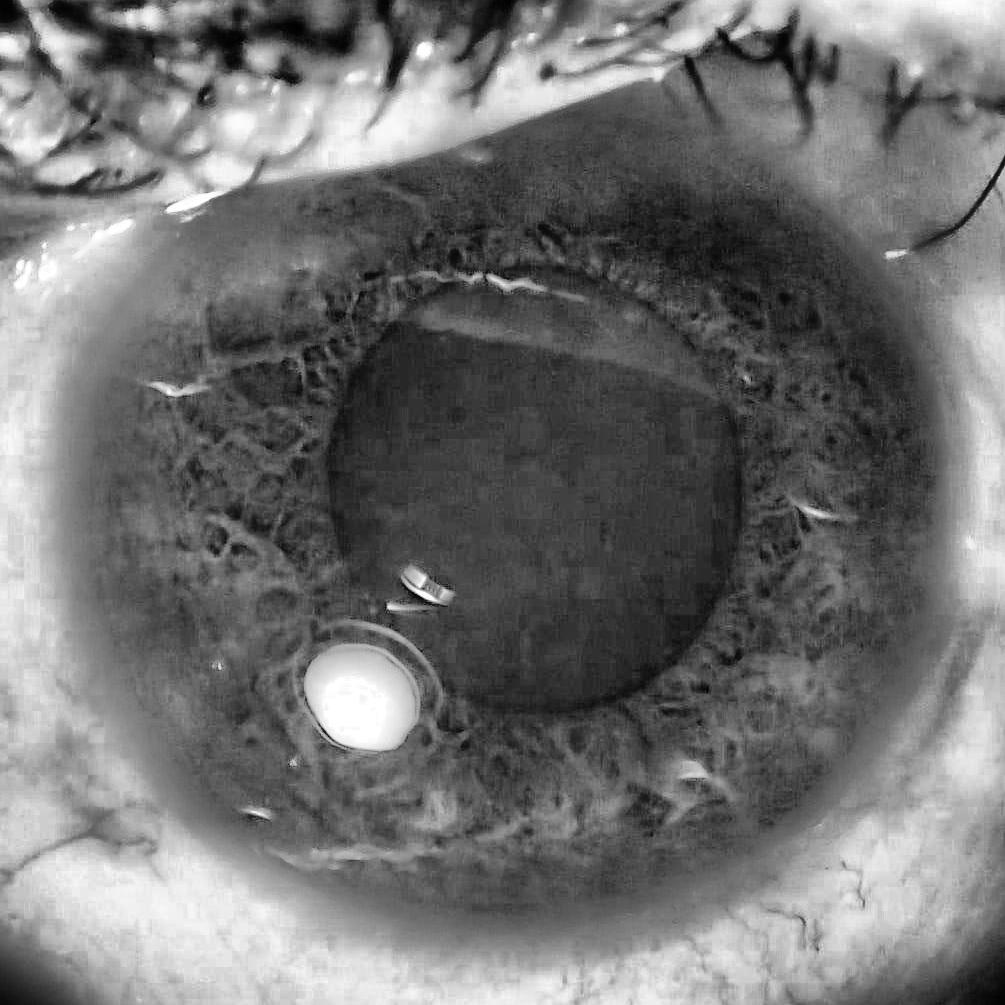}\\
	\includegraphics[width=0.16\textwidth]{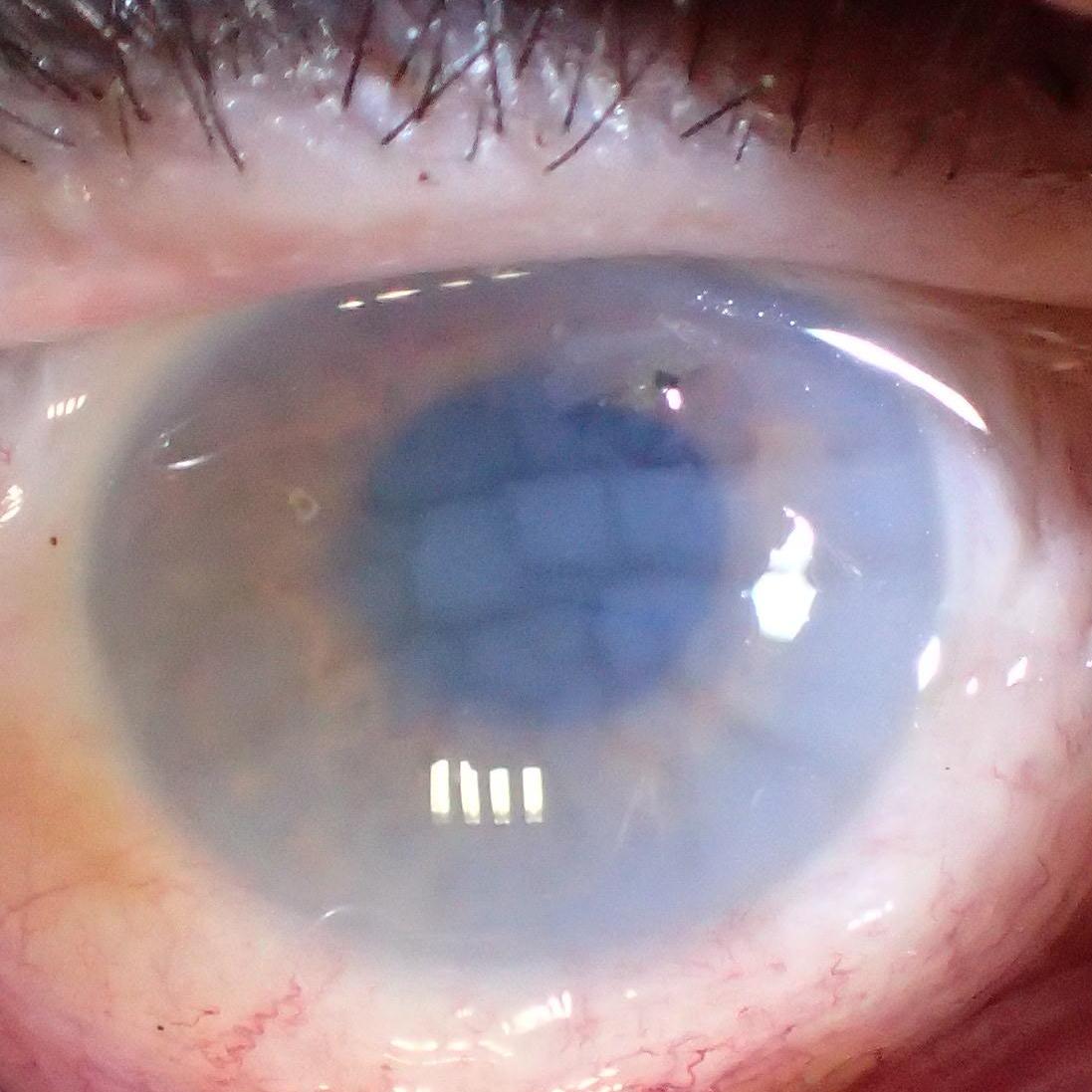}\hfill
	\includegraphics[width=0.16\textwidth]{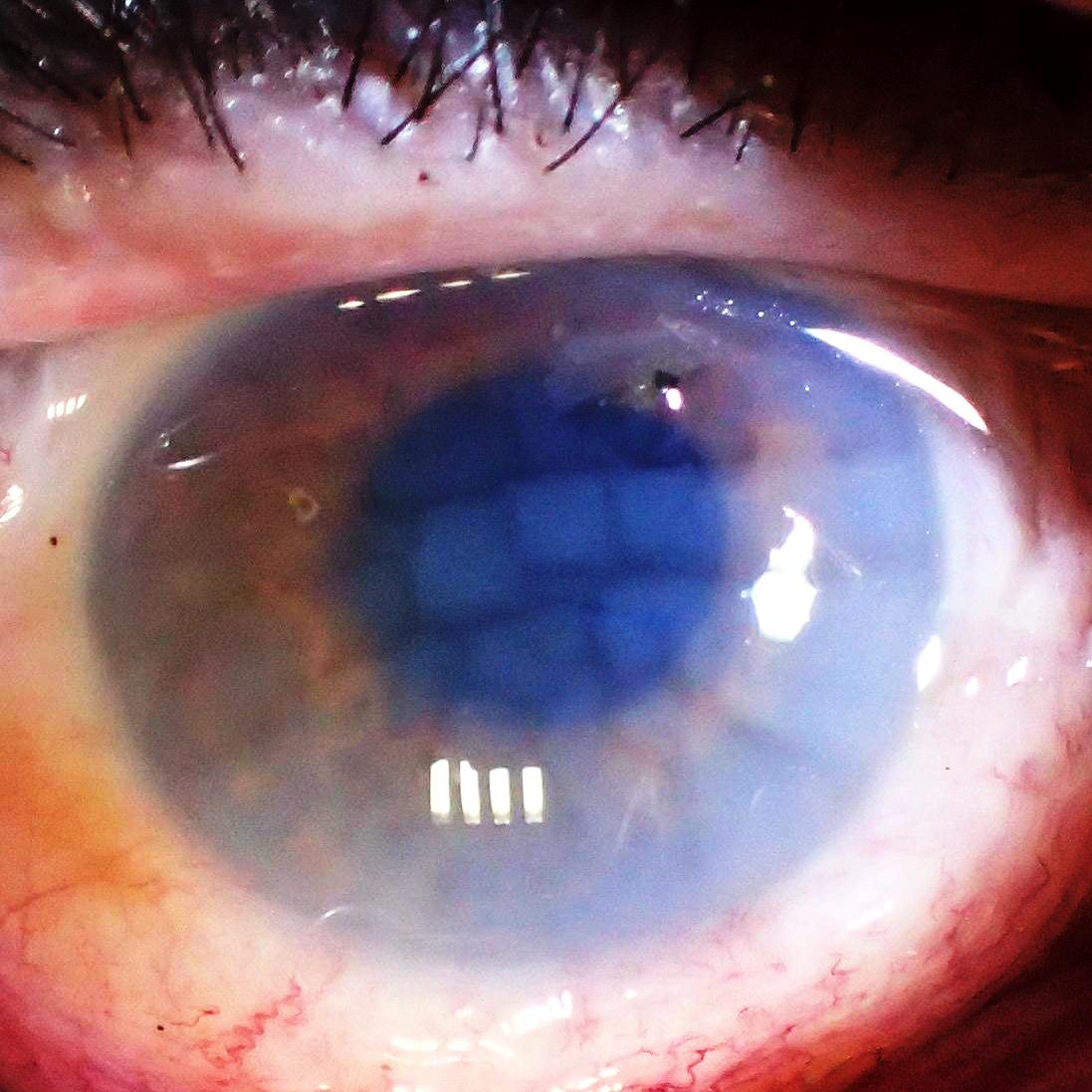}
	\includegraphics[width=0.16\textwidth]{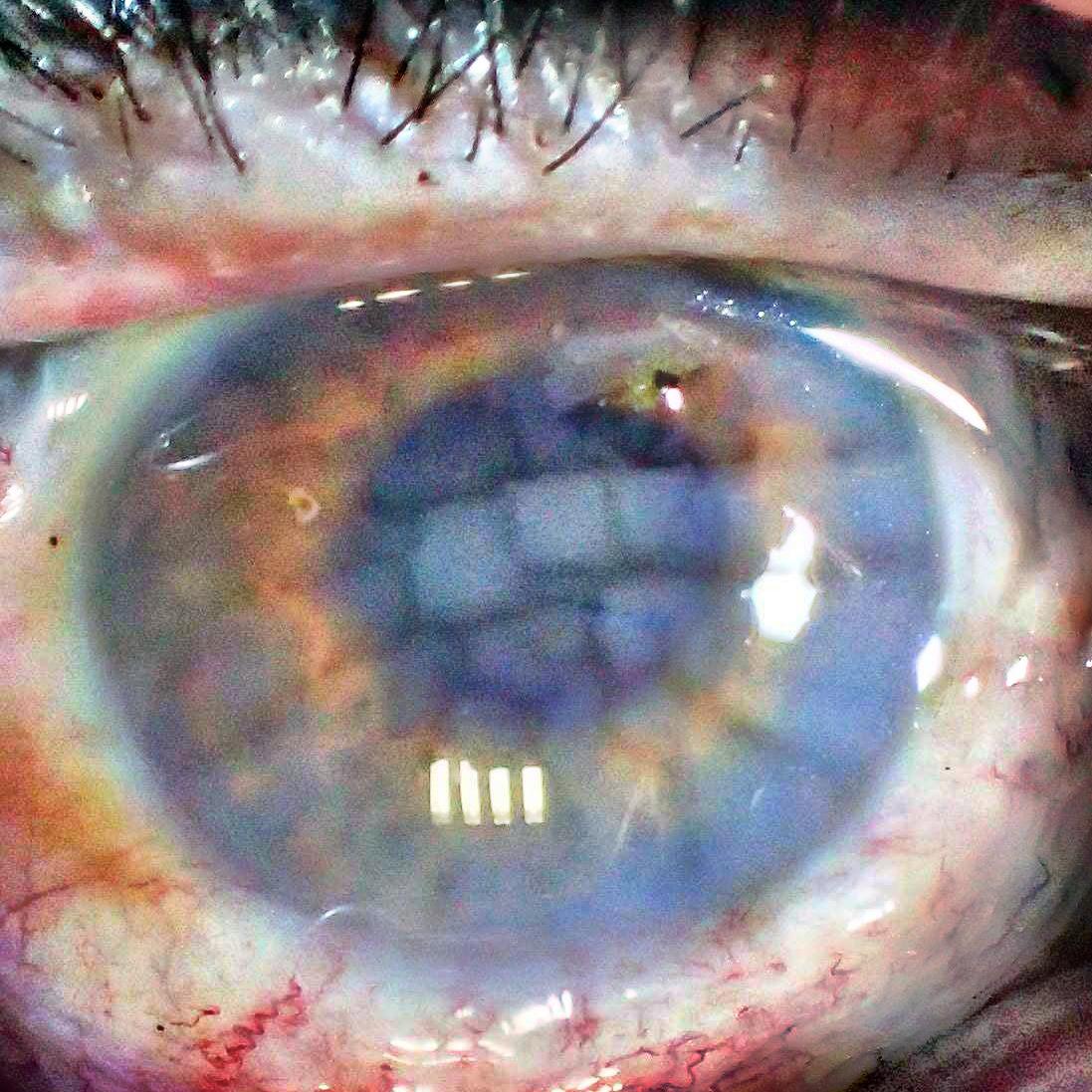}
	\includegraphics[width=0.16\textwidth]{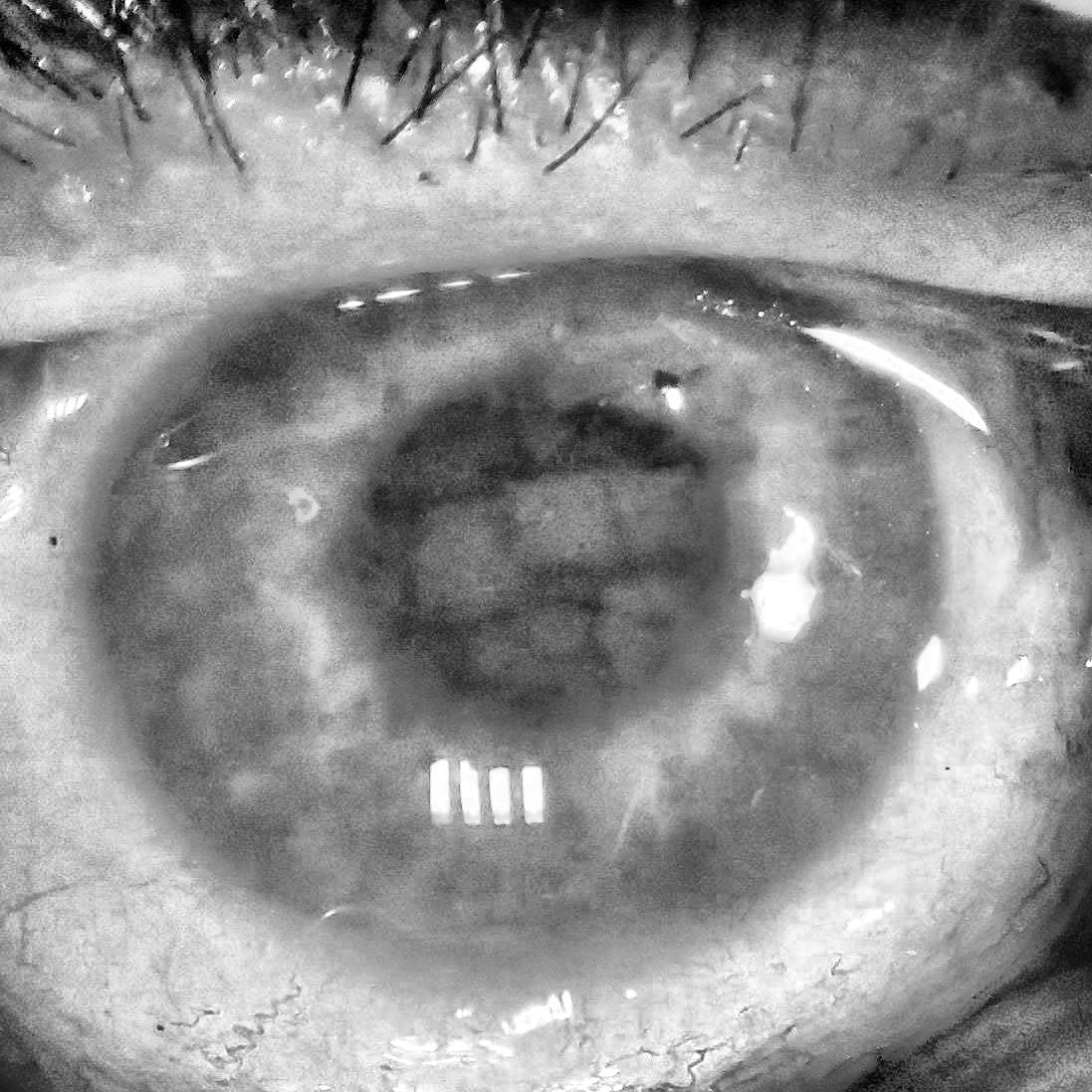}
	\includegraphics[width=0.16\textwidth]{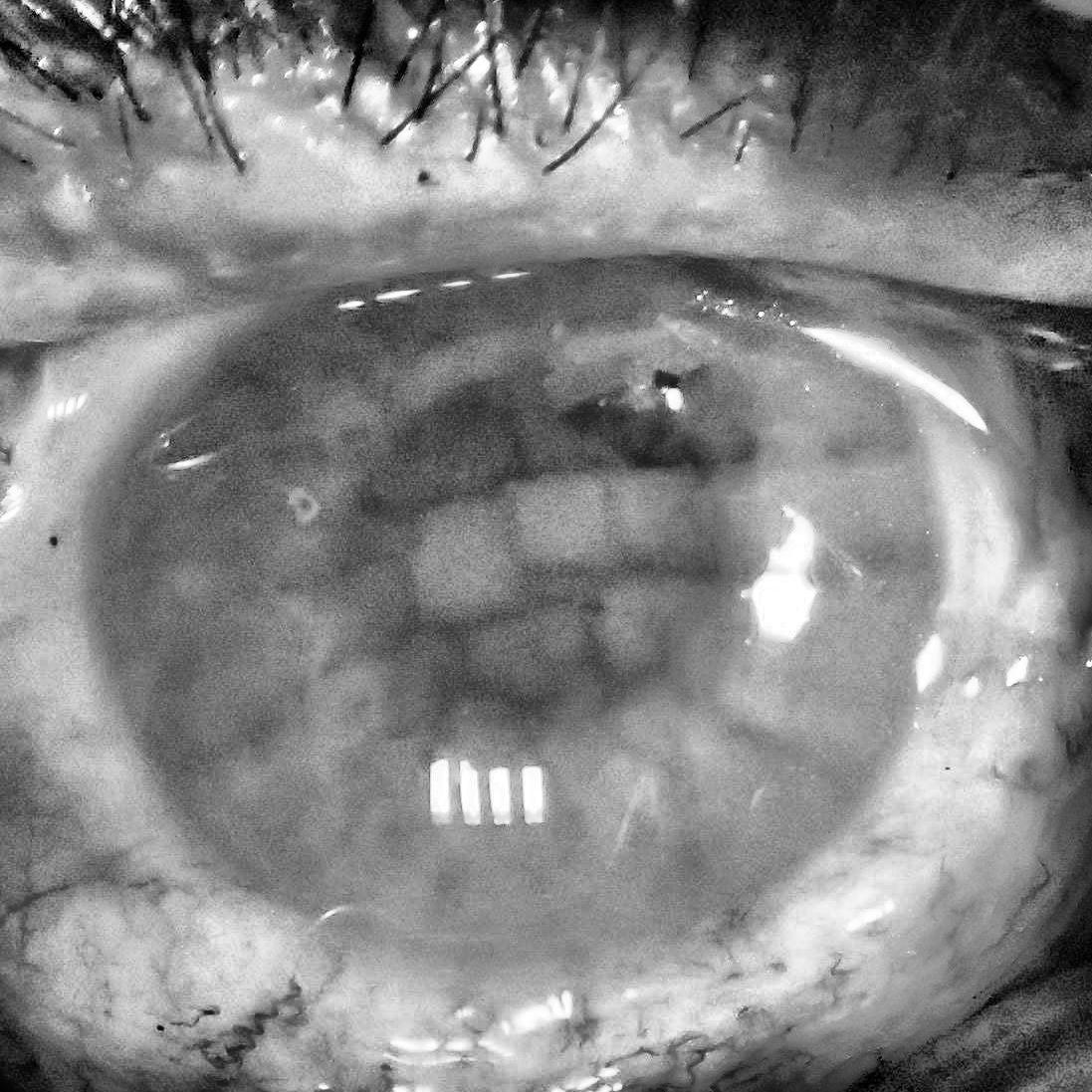}
	\includegraphics[width=0.16\textwidth]{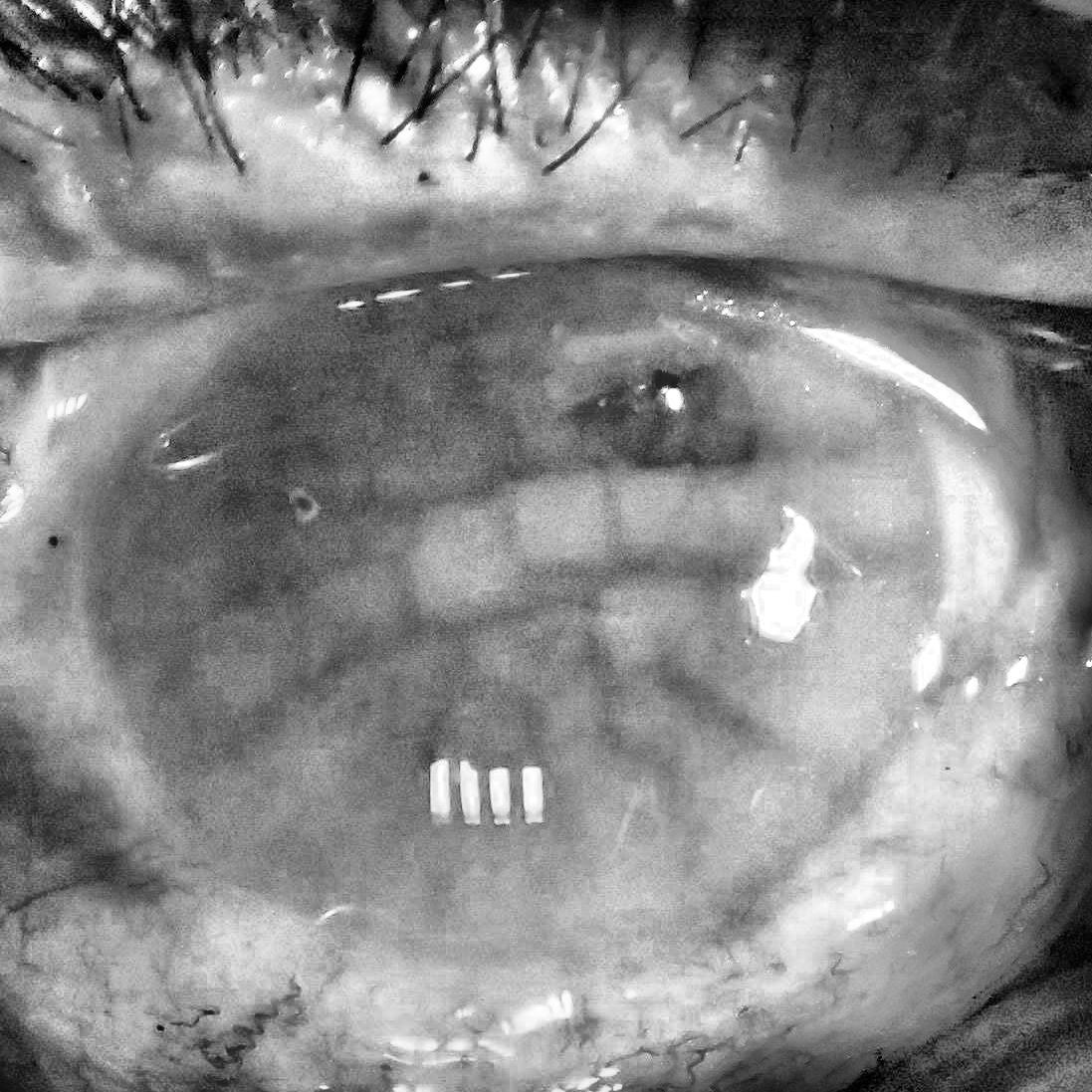}
	\caption{Image preprocessing applied to visible light images in order to boost iris texture visibility, from left to right: original image, histogram equalization, and CLAHE in RGB, red, green, and blue channel. Upper row: session 1. Bottom row: session 3.}
	\label{fig:VIS-preprocessing}
\end{figure*}

As for the visible light images, the histogram equalization and CLAHE normalization were applied to each of the RGB channels. Fig. \ref{fig:VIS-preprocessing} depicts these types of preprocessing applied to a sample collected in the first acquisition session, as well to a sample obtained in the third session. The RGB image made from three CLAHE-normalized RGB channels offer the best visibility of the iris texture. Notably, while in the first session image the individual iris features can be easily discerned in each of the three individual RGB channels, the blue channel provides almost no visibility of the iris texture in the third session, while showing the most of the corneal haze. Following this conclusion, we employ the CLAHE-normalized RGB image for the purpose of inspecting individual features. 
   
In contrast to NIR images, RGB ones seem to offer much worse visibility of the iris texture throughout the eight considered acquisition sessions. Especially, the opacity of the cornea can be observed as early as the third session, for which the NIR image still looked clear. The second problem is a large number of additional light reflections, which are invisible, or significantly reduced) in NIR illumination, which often cover the iris features. Out of four annotated features, only Feature 1 still resembles its original appearance in the final, borderline session. The other three are either covered by corneal opacity or by specular reflections. 

The close-up analysis provides the following observations and conclusions:
\begin{enumerate}
	\item \textbf{Feature decomposition rate.} Despite the well preserved overall post-mortem appearance of the iris, the individual features (such as crypts) vanish or change rather fast -- days instead of weeks. 
	\item \textbf{Illumination differences.} With NIR images, selected iris features could be identified even up to the borderline session, which was determined to be 407 hours; for visible light images this was also possible, but only for one out of four annotated features; a problem here are additional light reflections which make it difficult to obtained a good visibility of the iris texture, as well as corneal opacification, which is much more pronounced than in NIR images.
	\item \textbf{Image preprocessing.} Image preprocessing to boost contrast, such as histogram equalization, can help human examiners in analyzing individual iris features.
	\item \textbf{Examination equipment suggestions.} Visible light images offer better visibility of the iris texture, probably thanks to higher resolution output of the device (dSLR camera), albeit for a shorter post-mortem sample capture time horizons. NIR images, on the other hand, offer better overall iris visibility and chances for a longer preservation of the iris features. Our recommendation is to use a high-resolution camera operating in near-infrared, as typical VGA iris recognition cameras seem to produce images of insufficient quality for post-mortem evaluation. Also, using high f-stops (low aperture values) to get a sharp iris tissue image even with non-perfect focusing, and correct light source placement, so that it does not emphasize the corneal opacification can be beneficial for image quality and iris visibility.
\end{enumerate}

\begin{figure*}[h!]
	\centering 
	{\bf Feature annotation in RGB image:}\\\vskip1mm
	\includegraphics[width=0.12\textwidth]{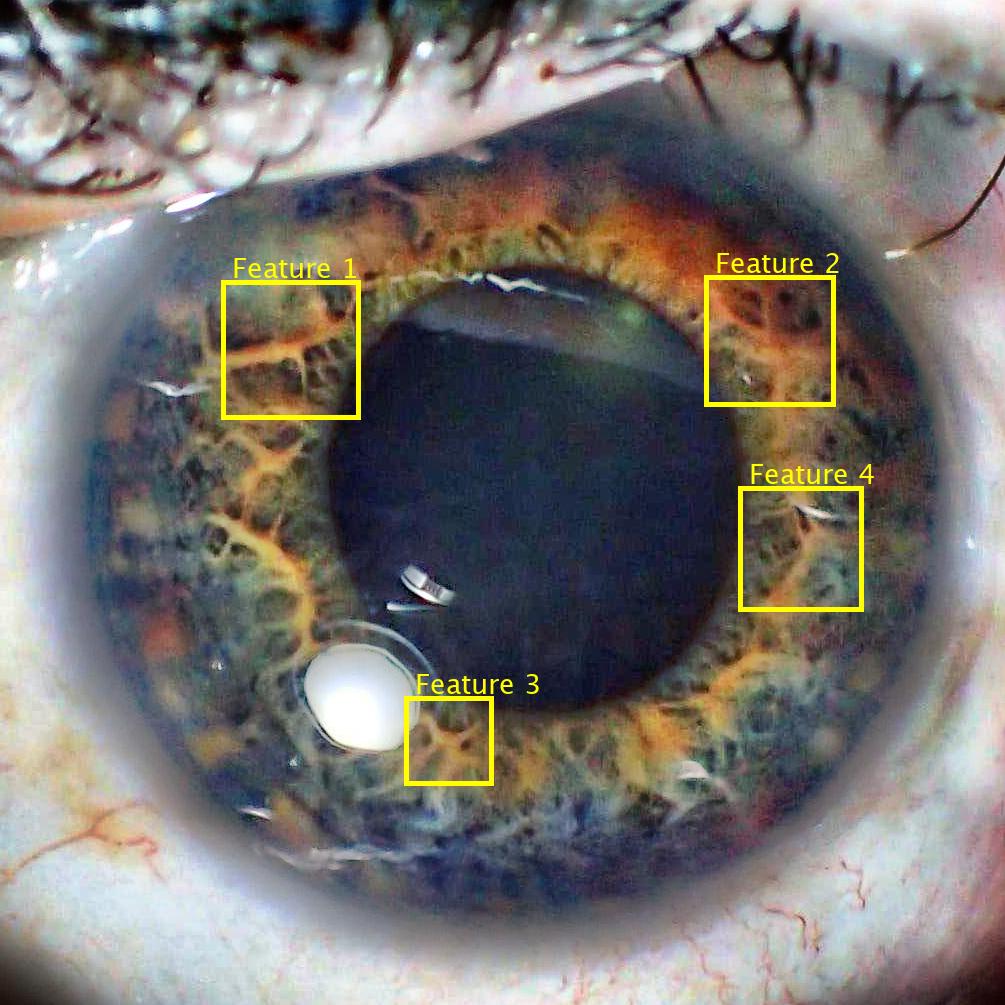}
	\includegraphics[width=0.12\textwidth]{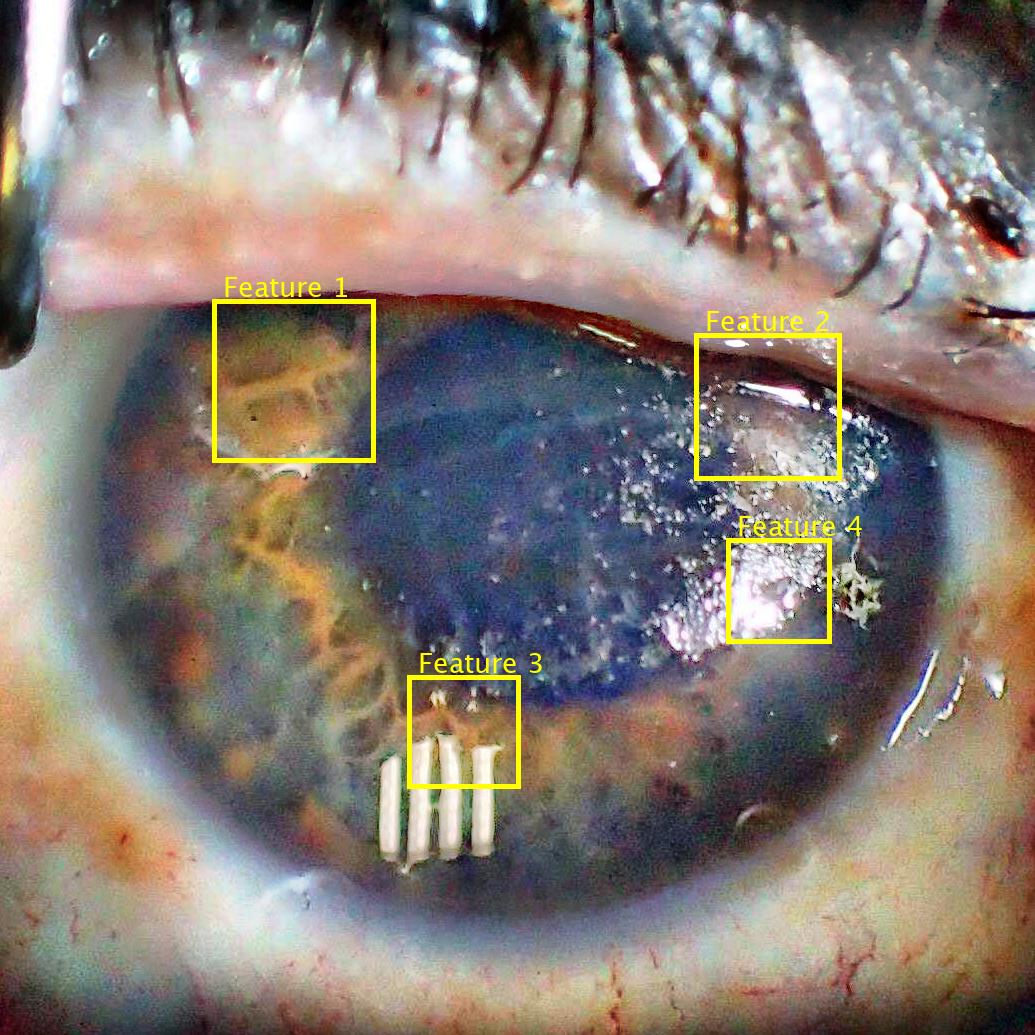}
	\includegraphics[width=0.12\textwidth]{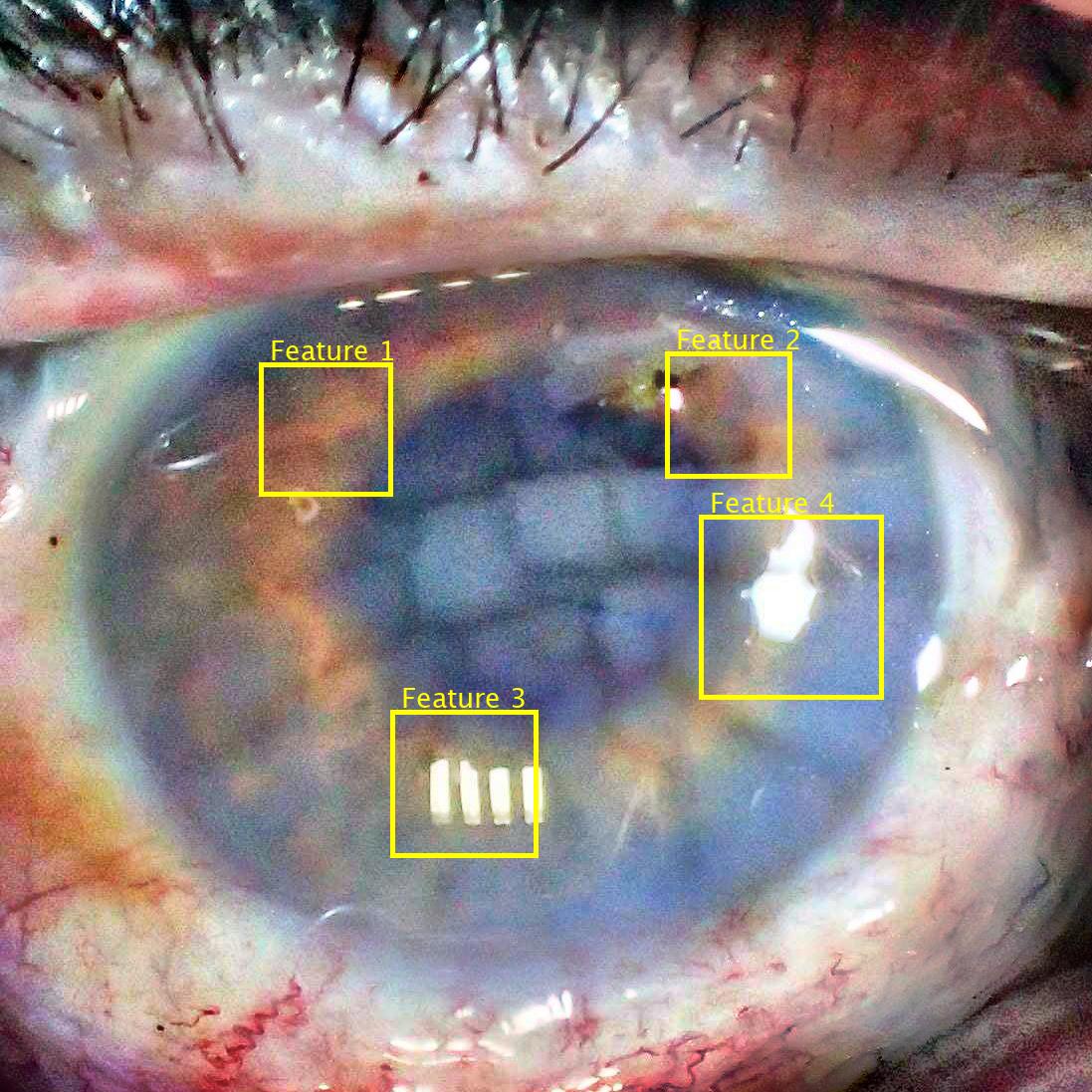}
	\includegraphics[width=0.12\textwidth]{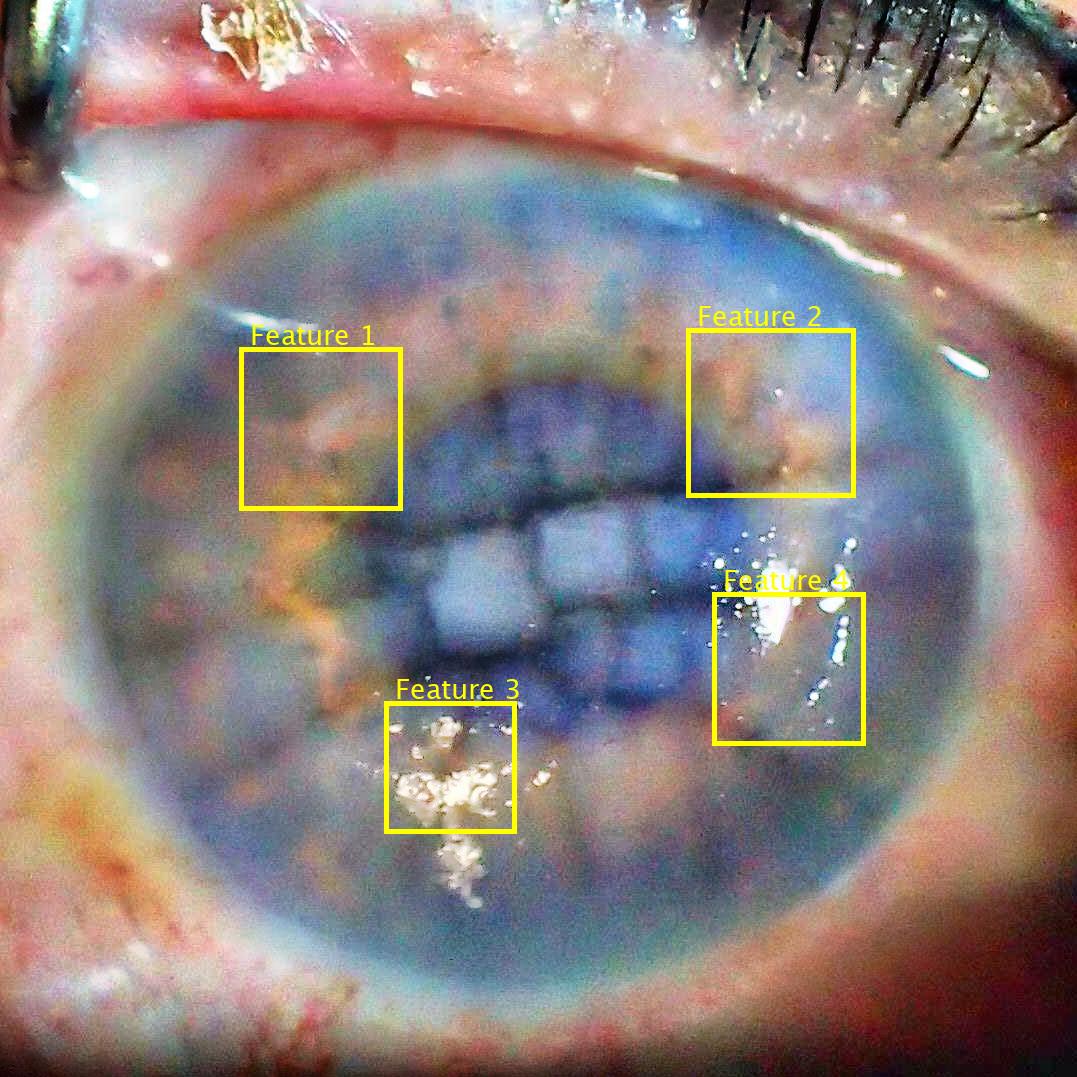}
	\includegraphics[width=0.12\textwidth]{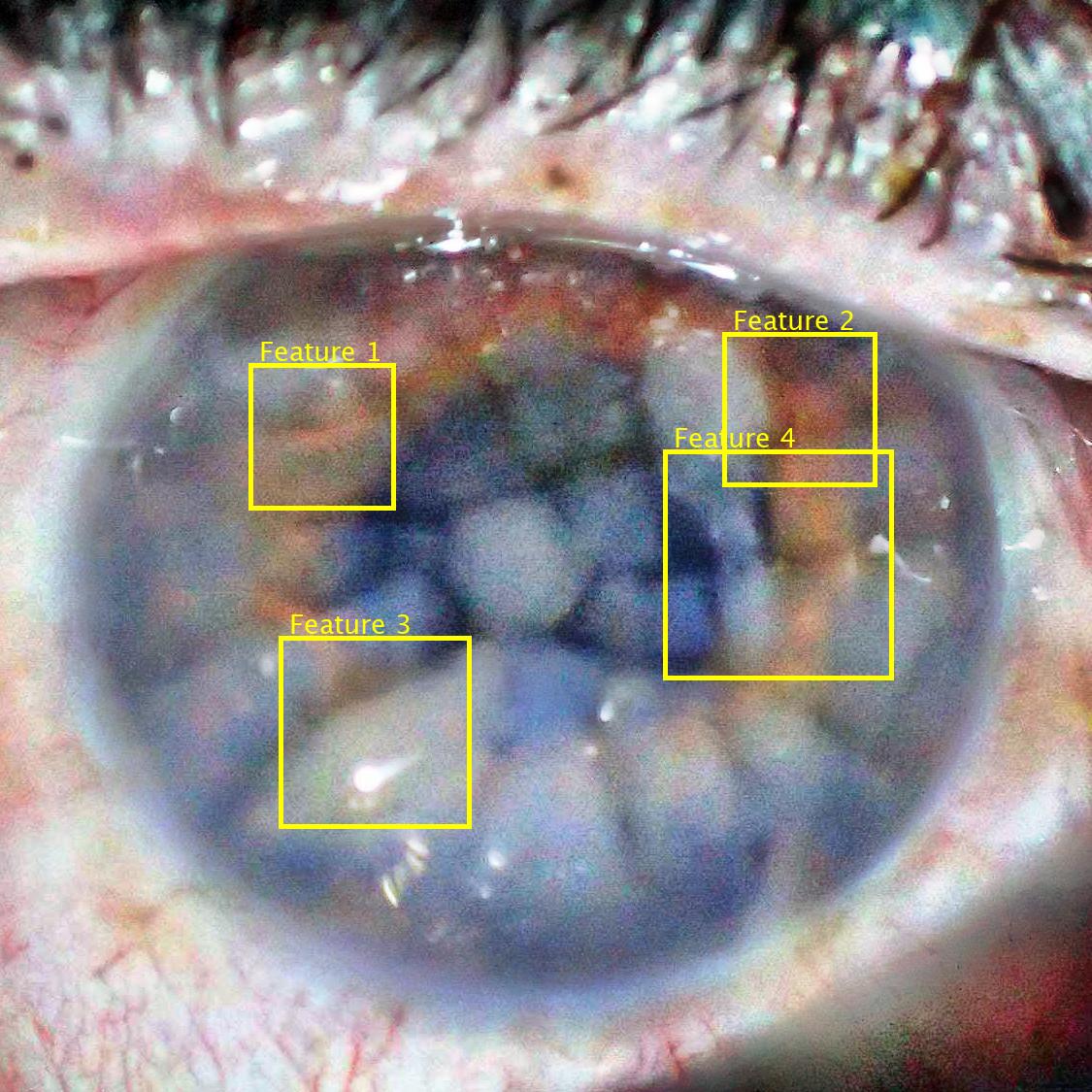}
	\includegraphics[width=0.12\textwidth]{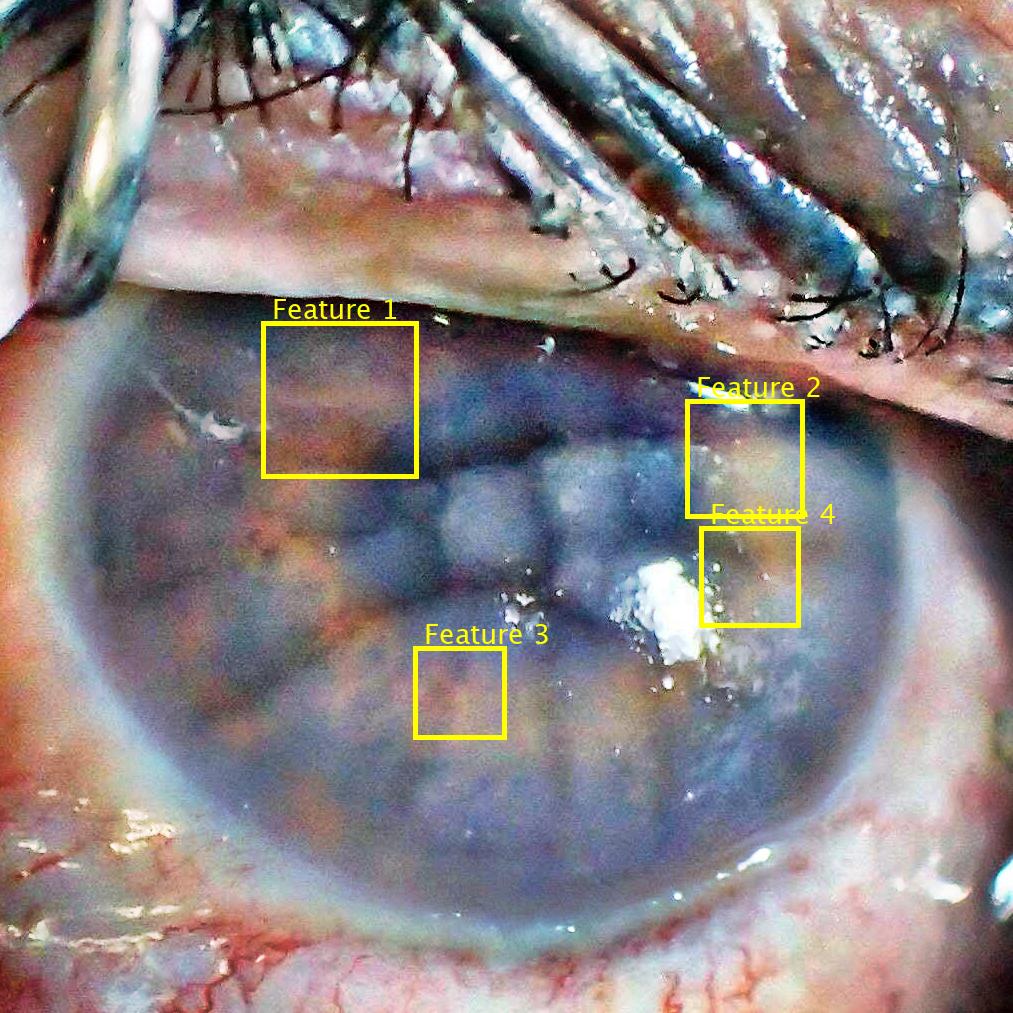}
	\includegraphics[width=0.12\textwidth]{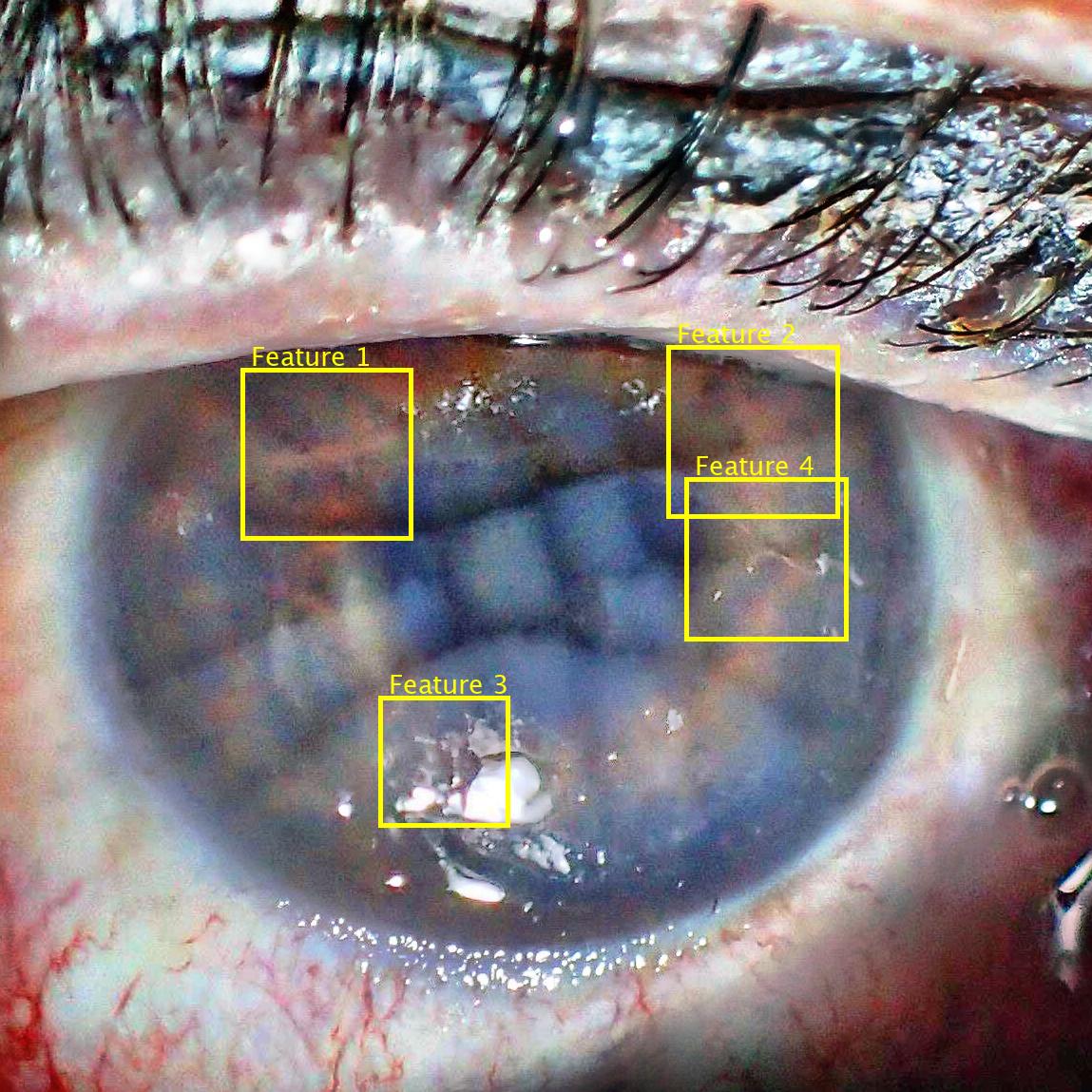}
	\includegraphics[width=0.12\textwidth]{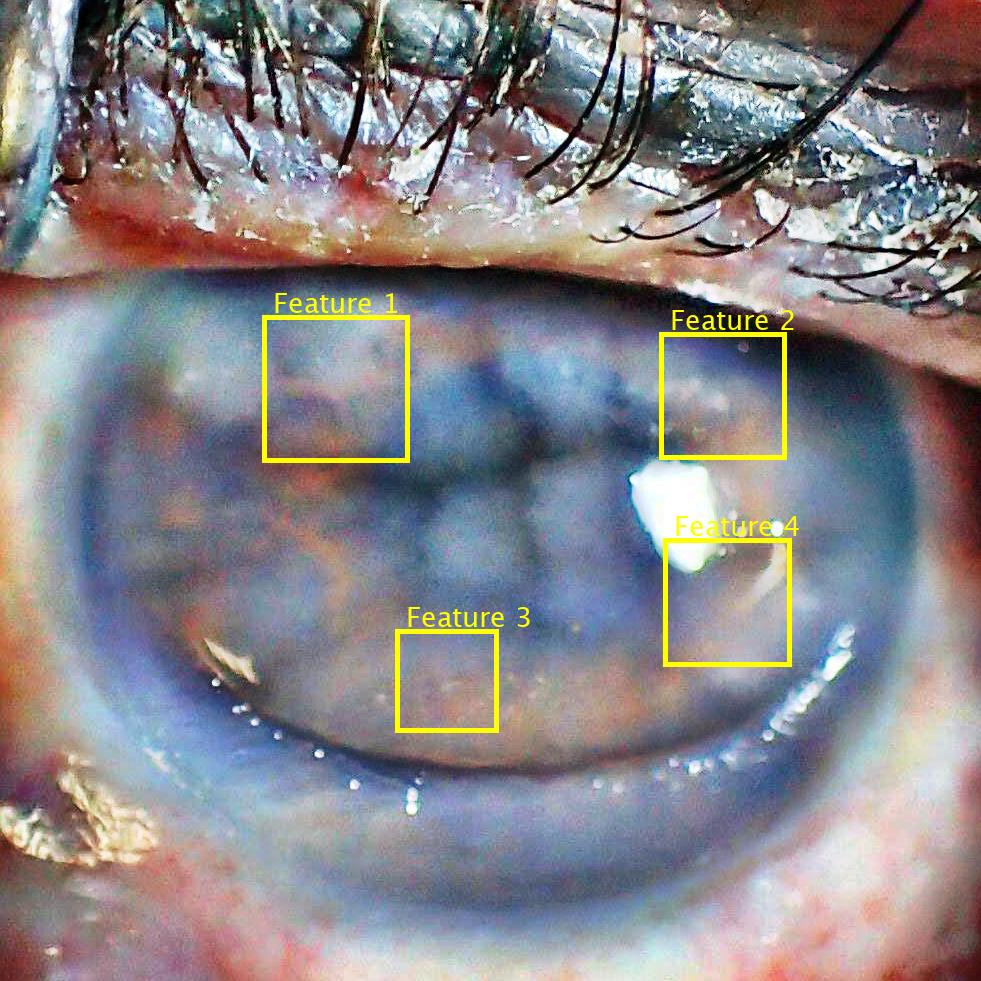}\\
	5 hours \hskip8mm 23 hours \hskip8mm 95 hours \hskip8mm 154 hours \hskip8mm 215 hours \hskip8mm 263 hours \hskip8mm 359 hours \hskip8mm 407 hours \\\vskip1mm
	{\bf Individual feature close-ups:}\\
	\hskip-107mm Feature 1:\\\vskip1mm
	\includegraphics[width=0.12\textwidth]{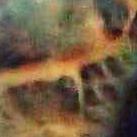}
	\includegraphics[width=0.12\textwidth]{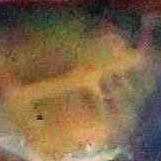}
	\includegraphics[width=0.12\textwidth]{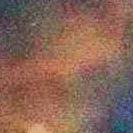}
	\includegraphics[width=0.12\textwidth]{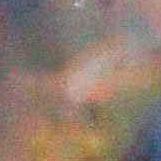}
	\includegraphics[width=0.12\textwidth]{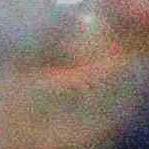}
	\includegraphics[width=0.12\textwidth]{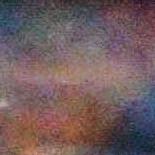}
	\includegraphics[width=0.12\textwidth]{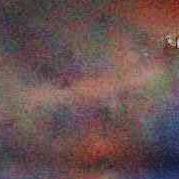}
	\includegraphics[width=0.12\textwidth]{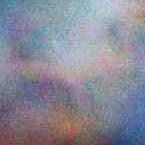}\\
	\hskip-107mm Feature 2:\\\vskip1mm
	\includegraphics[width=0.12\textwidth]{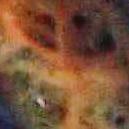}
	\includegraphics[width=0.12\textwidth]{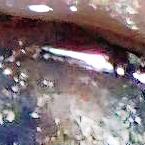}
	\includegraphics[width=0.12\textwidth]{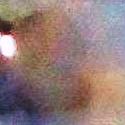}
	\includegraphics[width=0.12\textwidth]{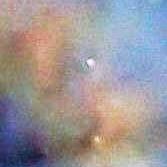}
	\includegraphics[width=0.12\textwidth]{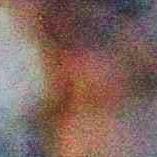}
	\includegraphics[width=0.12\textwidth]{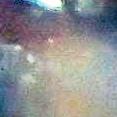}
	\includegraphics[width=0.12\textwidth]{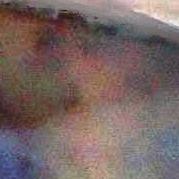}
	\includegraphics[width=0.12\textwidth]{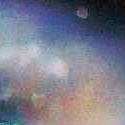}\\
	\hskip-107mm Feature 3:\\\vskip1mm
	\includegraphics[width=0.12\textwidth]{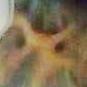}
	\includegraphics[width=0.12\textwidth]{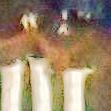}
	\includegraphics[width=0.12\textwidth]{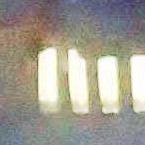}
	\includegraphics[width=0.12\textwidth]{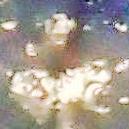}
	\includegraphics[width=0.12\textwidth]{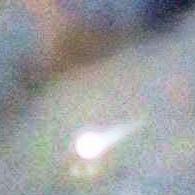}
	\includegraphics[width=0.12\textwidth]{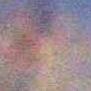}
	\includegraphics[width=0.12\textwidth]{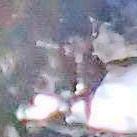}
	\includegraphics[width=0.12\textwidth]{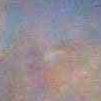}\\
	\hskip-107mm Feature 4:\\\vskip1mm
	\includegraphics[width=0.12\textwidth]{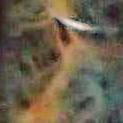}
	\includegraphics[width=0.12\textwidth]{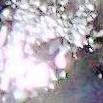}
	\includegraphics[width=0.12\textwidth]{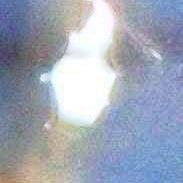}
	\includegraphics[width=0.12\textwidth]{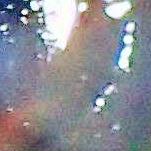}
	\includegraphics[width=0.12\textwidth]{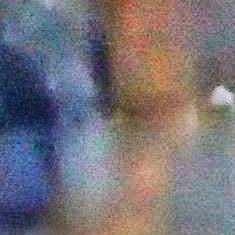}
	\includegraphics[width=0.12\textwidth]{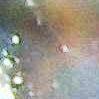}
	\includegraphics[width=0.12\textwidth]{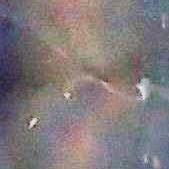}
	\includegraphics[width=0.12\textwidth]{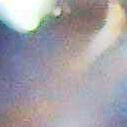}
	\caption{Same as in Fig. \ref{fig:closeup-NIR}, but done with RGB images.}
	\label{fig:closeup-RGB}
\end{figure*}

\section{Conclusions}
\label{sec:Conclusions}

This paper presents the first known to us detailed analysis of post-mortem changes observed in irises of cadavers kept in mortuary conditions from the biometric perspective. The first observation is that there are visible iris features possible to be matched by humans for a longer time (407 hours post-mortem in our example) than features extracted by off-the-shelf iris recognition matchers (occasional correct matches 120 hours post-mortem for all four matchers used in \cite{TrokielewiczPostMortemBTAS2016}, and on par with 407 hours post-mortem for the matcher offering the best accuracy).

The second observation from this analysis is that NIR illumination helps to better mitigate the corneal opacification when compared to visible-light acquisition, which suffers from very pronounced corneal opacity, as well as additional light reflections obstructing the view on iris tissue. Ideally, post-mortem-aware iris acquisition for automatic recognition purposes should thus incorporate NIR illumination and higher than typical ($640\times480$ pixels) image resolution. Using high f-stops is also suggested to ensure good focus and sharp iris image. 

This shows that iris can offer features useful in even a long-term post-mortem identification of humans, and may call for designing specialized feature extraction algorithms that would aid human examiners in their decisions.

\section*{Conflict of interest}
The authors declare that they have no conflict of interest.

\end{document}